\documentclass[10pt,twocolumn,letterpaper]{article}

\usepackage{cvpr}
\usepackage{times}
\usepackage{epsfig}
\usepackage{graphicx}
\usepackage{amsmath} 
\usepackage{amssymb}
\usepackage{bbm}

\usepackage{booktabs}
\usepackage[table]{xcolor}
\usepackage{multirow}
\usepackage{cite}

\cvprfinalcopy 

\usepackage{cleveref}[2012/02/15]
\crefformat{footnote}{#2\footnotemark[#1]#3}

\ifcvprfinal\fi

\begin{document}

\title{Unsupervised Category Discovery via Looped Deep Pseudo-Task Optimization\\ Using a Large Scale Radiology Image Database}

\author{Xiaosong Wang \ \ \ \ \ Le Lu \ \ \ \ \ Hoo-chang Shin \ \ \ \ \ Lauren Kim \ \ \ \ \ Isabella Nogues \ \ \ \ \ Jianhua Yao \\ Ronald Summers\\
	Imaging Biomarkers and Computer-aided Detection Laboratory \\
	Department of Radiology and Imaging Sciences\\
	National Institutes of Health Clinical Center\\
	10 Center Drive, Bethesda, MD 20892\\
	{\tt\small xiaosong.wang,le.lu,hoo-chang.shin,lauren.kim,isabella.nogues,rms@nih.gov, jyao@cc.nih.gov}
}

\maketitle

\begin{abstract}
	Obtaining semantic labels on a large scale radiology image database (215,786 key images from 61,845 unique patients) is a prerequisite yet bottleneck to train highly effective deep convolutional neural network (CNN) models for image recognition. Nevertheless, conventional methods for collecting image labels (e.g., Google search followed by crowd-sourcing) are not applicable due to the formidable difficulties of medical annotation tasks for those who are not clinically trained. This type of image labeling task remains non-trivial even for radiologists due to uncertainty and possible drastic inter-observer variation or inconsistency.
	
	~~~In this paper, we present a looped deep pseudo-task optimization procedure for automatic category discovery of visually coherent and clinically semantic (concept) clusters. Our system can be initialized by domain-specific (CNN trained on radiology images and text report derived labels) or generic (ImageNet based) CNN models. Afterwards, a sequence of pseudo-tasks are exploited by the looped deep image feature clustering (to refine image labels) and deep CNN training/classification using new labels (to obtain more task representative deep features). Our method is conceptually simple and based on the hypothesized  ``convergence'' of better labels leading to better trained CNN models which in turn feed more effective deep image features to facilitate more meaningful clustering/labels. We have empirically validated the convergence and demonstrated promising quantitative and qualitative results. Category labels of significantly higher quality than those in previous work are discovered. This allows for further investigation of the hierarchical semantic nature of the given large-scale radiology image database.
	
\end{abstract}

\section{Introduction} 

The rapid and tremendous success of applying deep convolutional neural networks (CNNs) \cite{krizhevsky2012imagenet,simonyan2014very,Szegedy2014Going} to many challenging computer vision tasks derives from the accessibility of the well-annotated ImageNet \cite{deng2009imagenet,russakovsky2014imagenet} and PASCAL VOC \cite{Everingham2015Pascal} datasets. Deep CNNs perform significantly better than previous shallow learning methods and hand-crafted image features, however, at the cost of requiring greater amounts of training data. 
ImageNet pre-trained deep CNN models \cite{jia2014caffe,krizhevsky2012imagenet,Lin2015Nin} serve an indispensable role to be bootstrapped upon for all externally-sourced data exploitation tasks \cite{Liang2015Baby,Chen2015Webly}. In the medical domain, however, no comparable labeled large-scale image dataset is available except the recent \cite{Shin2015}. Vast amounts of radiology images/reports are stored in many hospitals' Picture Archiving and Communication Systems (PACS), but the main challenge lies in how to obtain ImageNet-level semantic labels on a large collection of medical images \cite{Shin2015}. 

Nevertheless, conventional means of collecting image labels (e.g. Google image search using the terms from WordNet ontology hierarchy \cite{Miller1995}, SUN/PLACE databases \cite{Xiao2010Sun,Zhou2014Place} or NEIL knowledge base \cite{Chen2013Neil}; followed by crowd-sourcing \cite{deng2009imagenet}) are not applicable due to 1) the formidable difficulties of medical annotation tasks for clinically untrained annotators, 2) unavailability of a high quality or large capacity medical image search engine. On the other hand, even for well-trained radiologists, this type of ``assigning labels to images'' task is not aligned with their regular diagnostic routine work so that drastic inter-observer variations or inconsistency may be demonstrated. The protocols of defining image labels based on visible anatomic structures (often multiple), or pathological findings (possibly multiple) or using both cues have a lot of ambiguity. 

Shin et al. \cite{Shin2015} first extract the sentences depicting disease reference key images (similar concept to ``key frames in videos'') using natural language processing (NLP) out of $\sim780$K patients' radiology reports, and find 215,786 key images of 61,845 unique patients from PACS. Then, image categorization labels are mined via unsupervised hierarchical Bayesian document clustering, i.e. generative latent Dirichlet allocation (LDA) topic modeling \cite{blei2003latent}, to form 80 classes at the first level of hierarchy. The purely text-computed category information offers some coarse level of radiology semantics but is limited in two aspects: 1) The classes are {\em highly unbalanced}, in which one dominating category contains 113,037 images while other classes contain a few dozens. 2) The classes are not {\em visually} coherent. As a result, transfer learning from the CNN models trained in \cite{Shin2015} to other medical computer-aided detection (CAD) problems performs less compellingly than those transferred directly from ImageNet CNNs \cite{Shin2015Deep,krizhevsky2012imagenet,Szegedy2014Going}. 

In this paper, we present a {\bf L}ooped {\bf D}eep {\bf P}seudo-task {\bf O}ptimization (LDPO) approach for automatic category discovery of visually coherent and clinically semantic (concept) clusters. The true semantic category information is assumed to be latent and not directly observable. The main idea is to learn and train CNN models using pseudo-task labels (when human annotated labels are unavailable) and iterate this process with the expectation that pseudo-task labels will eventually resemble latent true image categories. Our work is partly related to the recent progress of semi-supervised learning or self-taught image classification, which has advanced both image classification and clustering processes~\cite{Singh2012DiscPat,Raina2007Self,Li2011Towards,Juneja2013Blocks,Dai2015EnProDeepFets,Dai2013EnPro}. The iterative optimization in \cite{Singh2012DiscPat,Juneja2013Blocks} seeks to identify discriminative local visual patterns and reject others, whereas our goal is to set better labels for all images during iterations towards auto-annotation. 

Our contributions are in several fold. \textbf{1)}, We propose a new ``iteratively updated'' deep CNN representation based on the LDPO technique. Thus it requires no hand-crafted image feature engineering \cite{Singh2012DiscPat,Raina2007Self,Li2011Towards,Juneja2013Blocks} which may be challenging for a large scale medical image database. Our method is conceptually simple and based on the hypothesized  ``convergence'' of better labels lead to better trained CNN models which in turn, offer more effective deep image features to facilitate more meaningful clustering/labels. {\it This looped property is unique to deep CNN classification-clustering models since other types of classifiers do not learn better image features simultaneously.} We use the database from \cite{Shin2015} to conduct experiments with the proposed method in different LDPO settings. Specifically, different pseudo-task initialization strategies, two CNN architectures of varying depths (i.e., AlexNet \cite{krizhevsky2012imagenet} and GoogLeNet~\cite{Szegedy2014Going}), different deep feature encoding schemes \cite{Cimpoi2015Filter,Cimpoi2015Deep} and clustering via K-means only or over-fragmented K-means followed by Regularized Information Maximization (RIM \cite{Gomes2010Discriminative} as an effective model selection method), are extensively explored and empirically evaluated.  \textbf{2)}, We consider the deep feature clustering followed by supervised CNN training as the outer loop and the deep feature clustering as the inner loop. Model selection on the number of clusters is critical and we carefully employ over-fragmented K-means followed by RIM model pruning/tuning to implement this criterion. This helps prevent cluster labeling amongst similar images, which can consequently compromise the CNN model training in the outer loop iteration. \textbf{3)}, The convergence of our LDPO framework can be observed and measured in both the cluster-similarity score plots and the CNN training classification accuracies. \textbf{4)}, Given the deep CNN LDPO models, hierarchical category relationships in a tree-like structure can be naturally formulated and computed from the final pairwise CNN classification confusion measures, as described in \ref{sec:hcl}. 
We will make our discovered image annotations (after reviewed and verified by board-certified radiologists in a with-humans-in-the-loop fashion \cite{Yu2015Construction}) together with trained CNN models publicly available upon publication.

To the best of our knowledge, this is the first work exploiting to integrate unsupervised deep feature clustering and supervised deep label classification for self-annotating a large scale radiology image database where the conventional means of image annotation are not feasible. The measurable LDPO ``convergence'' makes this ill-posed problem well constrained, at no human labeling costs. Our proposed LDPO method is also quantitatively validated using Texture-25 dataset \cite{Dai2015EnProDeepFets,Lazebnik2005Sparse} where the ``unsupervised'' classification accuracy improves over LDPO iterations. The ground truth labels of texture images \cite{Dai2015EnProDeepFets,Lazebnik2005Sparse} are known and used to measure the accuracy scores against LDPO clustering labels. Our results may grant the possibility of 1), investigating the hierarchical semantic nature (object/organ, pathology, scene, modality, etc.) of categories \cite{Rematas2015Dataset,Johnson2015Image}; 2), finer level image mining for tag-constrained object instance discovery and detection \cite{Wu2015Harvesting,Bazzani2015Self}, given the large-scale radiology image database. \vspace{-2mm}

\section{Related Work} 

{\bf Unsupervised and Semi-supervised Learning:} Dai \textit{et al.} \cite{Dai2015EnProDeepFets,Dai2013EnPro} study the semi-supervised image classification/clustering problem on texture \cite{Lazebnik2005Sparse}, small to middle-scale object classes (e.g., Caltech-101 \cite{FeiFei2004101}) and scene recognition datasets \cite{Quattoni2009indoor}. By exploiting the data distribution patterns that are encoded by so called ensemble projection (EP) on a rich set of visual prototypes, the new image representation derived from clustering is learned for recognition. Graph based approaches \cite{Liu2010Large,Kingma2014SSL} are used to link the unlabeled image instances to labeled ones as anchors and propagate labels by exploiting the graph topology and connectiveness weights. In an unsupervised manner, Coates \textit{et al.} \cite{Coates2011} employ k-means to mine image patch filters and then utilize the resulted filters for feature computation. Surrogate classes are obtained by augmenting each image patch with its geometrically transformed versions and a CNN is trained on top of these surrogate classes to generate features~\cite{Dosovitskiy2014}. Wang \textit{et al.} \cite{Wang2015Unsupervised} design a Siamese-triplet CNN network, leveraging object tracking information in $100$K unlabeled videos to provide the supervision for visual representation learning. Our work initializes an unlabeled image collection with labels from a pseudo-task (e.g., text topic modeling generated labels \cite{Shin2015}) and update the labels through an iterative looped optimization of deep CNN feature clustering and CNN model training (towards better deep image features). 

{\bf Text and Image:} \cite{berg2013babytalk} is a seminal work that models the semantic connections between image contents and the text sentences. Those texts describe cues of detecting objects of interest, attributes and prepositions and can be applied as contextual regularizations. \cite{Karpathy2015Deep} proposes a structured objective to align the CNN based image region descriptors and bidirectional Recurrent Neural Networks (RNN) over sentences through the multimodal embedding. \cite{Vinyals2015Show} presents a deep recurrent architecture from ``Sequence to Sequence'' machine translation \cite{Sutskever2014Sequence} to generate image description in natural sentences, via maximizing the likelihood of the target description sentence given the training image.  \cite{Sun2015Automatic} applies extensive NLP parsing techniques (e.g., unigram terms and grammatical relations) to extract concepts that are consequently filtered by the discriminative power of visual cues and grouped by joint visual and semantic similarities. \cite{Chen2015Sense} further investigates an image/text co-clustering framework to disambiguate the multiple semantic senses of some Polysemy words. The NLP parsing in radiology reports is arguably much harder than processing those public datasets of image captions  \cite{Karpathy2015Deep,Vinyals2015Show,berg2013babytalk} where most plain text descriptions are provided. Radiologists often rule out or indicate pathology/disease terms, {\em not existing in the corresponding key images}, but based on patient priors and other long-range contexts or abstractions. In \cite{Shin2015Interleaved}, only $\sim8$\% key images (18K out of 216K) can be tagged from NLP with the moderate confidence levels. We exploit the interactions from the text-derived image labels, to the proposed LDPO (mainly operating in the image modality) and the final term extraction from image groups.

{\bf Domain Transfer and Auto-annotation:} Deep CNN representation has made transfer learning or domain adaption among different image datasets practical, via straightforward fine-tuning \cite{Girshick2015RCNN,Razavian2014CNN}. Using pre-trained deep CNNs allows for the cross-domain transfer between weakly supervised video labels and noisy image labels. It can further output localized action frames by mutually filtering out low CNN-confidence instances \cite{Sun2015Temporal}. A novel CNN architecture is exploited for deep domain transfer to handle unlabeled and sparsely labeled target domain data \cite{Tzeng2015Simultaneous}. An image label auto-annotation approach is addressed via multiple instance learning \cite{Wu2015Deep} but the target domain is restricted to a small subset (25 out of 1000 classes) of ImageNet \cite{deng2009imagenet} and SUN \cite{Xiao2010Sun}. \cite{Wigness2015} introduces a method to identify a hierarchical set of unlabeled data clusters (spanning a spectrum of visual concept granularities) that are efficiently labeled to produce high performing classifiers (thus less label noise at instance level). By learning visually coherent and class balanced labels through LDPO, we expect that the studied large-scale radiology image database can markedly improve its feasibility in domain transfer to specific CAD problems where very limited training data are available per task.

\section{Looped Deep Pseudo-Task Optimization}\label{sec-method}

Traditional detection and classification problems in medical imaging, e.g. Computer Aided Detection (CAD) \cite{roth2015improving}, require precise labels of lesions or diseases as the training/testing ground-truth. This usually requires a large amount of annotation from well-trained medical professionals (especially at the era of ``deep learning''). Employing and converting the medical records stored in the PACS into labels or tags is very challenging \cite{Shin2015Interleaved}. Our approach performs the category discovery in an empirical manner and returns accurate key-word category labels for all images, through an iterative framework of deep feature extraction, clustering, and deep CNN model fine-tuning. 

\begin{figure*}
	\begin{center}
		\includegraphics[width=1.0\linewidth]{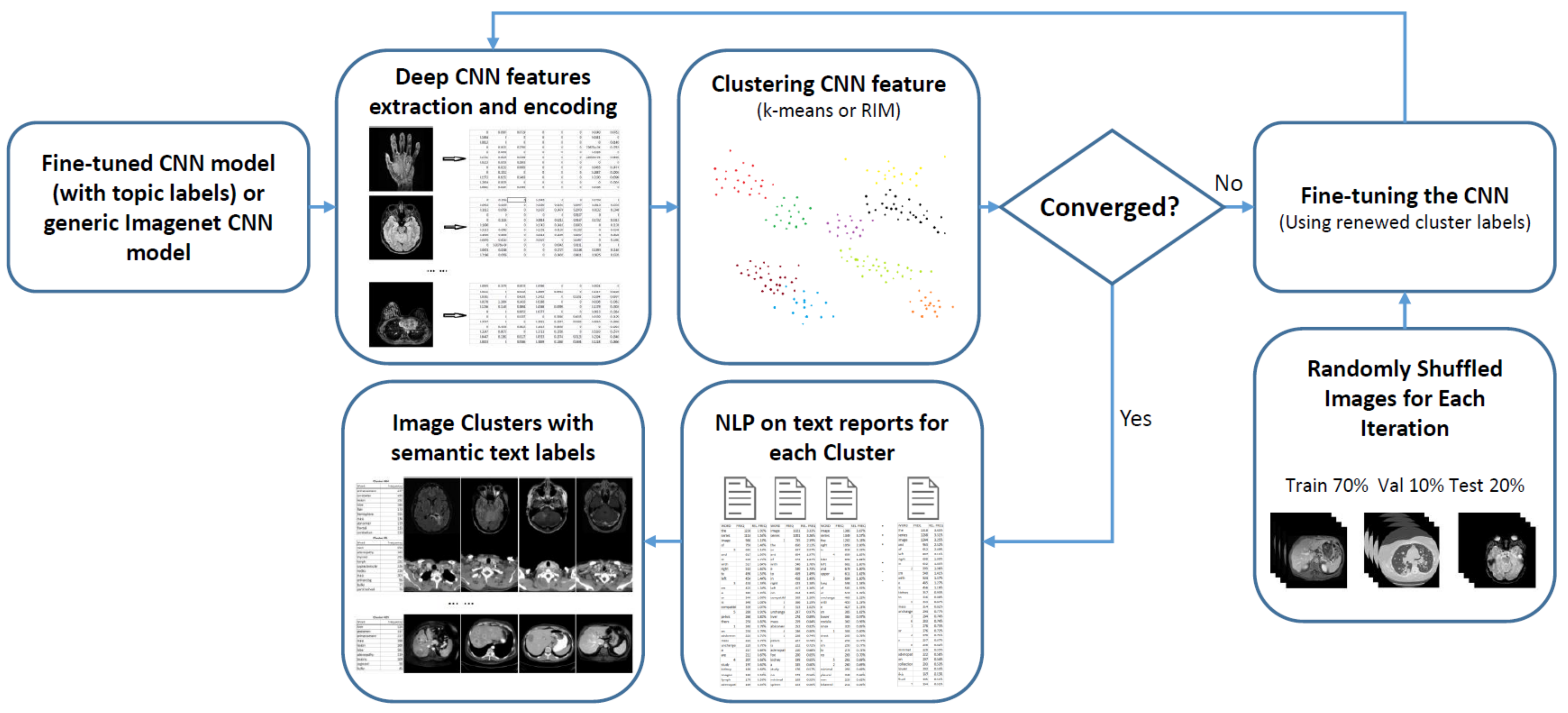}
	\end{center}
	\caption{The overview of looped deep pseudo-task optimization framework.}
	\label{fig:flowchart:png} 
\end{figure*}

As illustrated in Fig.~\ref{fig:flowchart:png}, the iterative process begins by extracting the deep CNN feature based on either a fine-tuned (with high-uncertainty radiological topic labels \cite{Shin2015}) or generic (from ImageNet labels \cite{krizhevsky2012imagenet}) CNN model. Next, the deep feature clustering with $k$-means or $k$-means followed by RIM is exploited. By evaluating the purity and mutual information between discovered clusters, the system either terminates the current iteration (which leads to an optimized clustering output) or takes the refined cluster labels as the input to fine-tune the CNN model for the following iteration. Once the visually coherent image clusters are obtained, the system further extracts semantically meaningful text words for each cluster. All corresponding patient reports per category cluster are finally adopted for the NLP. Furthermore, the hierarchical category relationship is built using the class confusion measures of the latest converged CNN classification models. 

\subsection{Convolution Neural Networks}

The proposed LDPO framework is applicable to a variety of CNN models. We analyze the CNN activations from layers of different depths in AlexNet~\cite{krizhevsky2012imagenet} and GoogLeNet~\cite{Szegedy2014Going}. Pre-trained models on the ImageNet ILSVRC data are obtained from Caffe Model Zoo \cite{jia2014caffe}. We also employ the Caffe CNN implementation~\cite{jia2014caffe} to perform fine-tuning on pre-trained CNNs using the key image database (from \cite{Shin2015}). Both CNN models with/without fine-tuning are used to initialize the looped optimization. AlexNet is a common CNN architecture with 7 layers and the extracted features from its convolutional or fully-connected layers have been broadly investigated \cite{Girshick2015RCNN,Razavian2014CNN,Karpathy2015Deep}. The encoded convolutional features for image retrieval tasks are introduced in \cite{Ng15}, which verifies the image representation power of convolutional features. In our experiments we adopt feature activations of both the 5th convolutional layer $Conv5$ and 7th fully-connected (FC) layer $FC7$ as suggested in~\cite{Cimpoi2015Deep,Chatfield14}. GoogLeNet is a much deeper CNN architecture compared to AlexNet, which comprises 9 inception modules and an average pooling layer. Each inception modules is truly a set of convolutional layers with multiple window sizes, i.e. $1\times1, 3\times3, 5\times5$. Similarly, we explore the deep image features from the last inception layer $Inception5b$ and final pooling layer $Pool5$. Table~\ref{tab:model} illustrates the detailed model layers and their activation dimensions. 
\small{
	\begin{table}[t]
		\caption{Configurations of CNN output layers and encoding methods (Output dimension is 4096, except the last row as 1024).}
		\centering
		\begin{tabular}{|l||c|c|c|}
			\hline\hline \small 
			{\bf CNN model}   & {\bf Layer} & {\bf Activations} & {\bf Encoding} \\
			\hline\hline
			{\bf AlexNet}    & Conv5 & $(13,13,256)$ & FV+PCA \\
			\hline
			{\bf AlexNet}   & Conv5 & $(13,13,256)$ & VLAD+PCA \\
			\hline
			{\bf AlexNet}   & FC7 & 4096 & $-$ \\ 
			\hline
			{\bf GoogLeNet}    & Inc.5b & $(7,7,1024)$ & VLAD+PCA \\ 
			\hline
			{\bf GoogLeNet}   & Pool5 & 1024 & $-$ \\
			\hline
		\end{tabular}\label{tab:model}
	\end{table}
}

\subsection{Encoding Images using Deep CNN Features} 

While the features extracted from fully-connected layer are able to capture the overall layout of objects inside the image, features computed at the last convolution layer preserve the local activations of images. Different from the standard max-pooling before feeding the fully-connected layer, we adopt the same setting (\cite{Cimpoi2015Filter}) to encode the convolutional layer outputs in a form of dense pooling via Fisher Vector (FV)~\cite{Perronnin2010FV} and Vector Locally Aggregated Descriptor (VLAD)~\cite{Jegou2012VLAD}. Nevertheless, the dimensions of encoded features are much higher than those of the FC feature. Since there is redundant information from the encoded features and we intend to make the results comparable between different encoding schemes, Principal Component Analysis (PCA) is performed to reduce the dimensionality to 4096, equivalent to the FC features' dimension.     

Given a pre-trained (generic or domain-specific) CNN model (i.e., Alexnet or GoogLeNet), an input image $I$ is resized to fit the model definition and feed into the CNN model to extract features $\{f^{L}_{i,j}\}$ ($1\leqslant i,j\leqslant s^{L}$) from the $L$-th convolutional layer with dimensions $s^{L}\times s^{L}\times d^{L}$, e.g., $13\times13\times256$ of $Conv5$ in AlexNet and $7\times7\times1024$ of $Pool5$ in GoogLeNet. For Fisher Vector implementation, we use the settings as suggested in~\cite{Cimpoi2015Deep}: 64 Gaussian components are adopted to train the Gaussian mixture Model(GMM). The dimension of resulted FV features is significantly higher than $FC7$'s, i.e. $32768 (2\times64\times256) \ vs \ 4096$. After PCA, the FV representation per image is reduced to a $4096$-component vector. A list of deep image features, the encoding methods and output dimensions are provided in Table~\ref{tab:model}. To be consistent with the settings of FV representation, we initialize the VLAD encoding of convolutional image features by $k$-means clustering with $k=64$. Thus the dimensions of VLAD descriptors are $16384(64\times256)$ of $Conv5$ in AlexNet and $65536(64\times1024)$ of $Inception5b$ in GoogLeNet. PCA further reduces the dimensions of both to $4096$. 

\subsection{Image Clustering} 

Image clustering plays an indispensable role in our LDPO framework. We hypothesize that the newly generated clusters driven by looped pseudo-task optimization are better than the previous ones in the following terms: 1) Images in each cluster are visually more coherent and discriminative from instances in other clusters; 2) The numbers of images per cluster are approximately equivalent to achieve class balance; 3) The number of clusters is self-adaptive according to the statistical properties of a large collection of image data. Two clustering methods are employed here, i.e. $k$-means alone and an over-segmented $k$-means (where $K$ is much larger than the first setting, e.g., 1000) followed by Regularized Information Maximization (RIM)~\cite{Gomes2010Discriminative} for model selection and optimization.

$k$-means is an efficient clustering algorithm provided that the number of clusters is known. We explore $k$-means clustering here for two reasons: 1) To set up the baseline performance of clustering on deep CNN image features by fixing the number of clusters $k$ at each iteration; 2) To initialize the RIM clustering since $k$-means is only capable of fulfilling our first two hypotheses, and RIM will help satisfy the third. Unlike $k$-means, RIM works with fewer assumptions on the data and categories, e.g. the number of clusters. It is designed for discriminative clustering by maximizing the mutual information between data and the resulted categories via a complexity regularization term. The objective function is defined as 
\begin{equation}
f(\mathbf{W};\mathbf{F},\lambda)=I_{\mathbf{W}}\{c;\mathbf{f}\}-R(\mathbf{W};\lambda),
\label{eq:RIM:objective}
\end{equation}
where $c\in\{1,...,K\}$ is a category label, $\mathbf{F}$ is the set of image features $\mathbf{f_{i}}=(f_{i1},...,f_{iD})^{T}\in\mathbb{R}^{D}$. $I_{\mathbf{W}}\{c;\mathbf{f}\}$ is an estimation of the mutual information between the feature vector $\mathbf{f}$ and the label $c$ under the conditional model $p(c|\mathbf{f},\mathbf{W})$. $R(\mathbf{W};\lambda)$ is the complexity penalty and specified according to $p(c|\mathbf{f},\mathbf{W})$. As demonstrated in~\cite{Gomes2010Discriminative}, we adopt the unsupervised multilogit regression cost. The conditional model and the regularization term are consequently defined as
\begin{eqnarray}
p(c=k|\mathbf{f},\mathbf{W})&\propto& exp(w^{T}_{k}\mathbf{f}+b_{k}) \\  R(\mathbf{W};\lambda)&=&\lambda\sum_{k}w^{T}_{k}w_{k},
\label{eq:RIM:multilogit}
\end{eqnarray}
where $\mathbf{W}=\{\mathbf{w}_{1},...,\mathbf{w}_{K},b_{1},...,b_{K}\}$ is the set of parameters and $\mathbf{w}_{k}\in\mathbb{R}^{D},b_{k}\in\mathbb{R}$. Maximizing the objective function is now equivalent to solving a logistic regression problem. $R$ is the $L_{2}$ regulator of weight $\{w_{k}\}$ and its power is controlled by $\lambda$. Large $\lambda$ values will enforce to reduce the total number of categories considering that no penalty is given for unpopulated categories~\cite{Gomes2010Discriminative}. This characteristic enables RIM to attain the optimal number of categories coherent to the data. $\lambda$ is fixed to $1$ in all our experiment.  

\subsection{Convergence in Clustering and Classification} \label{sec:cl} 

Before exporting the newly generated cluster labels to fine-tune the CNN model of the next iteration, the LDPO framework will evaluate the quality of clustering to decide if convergence has been achieved. Two convergence measurements have been adopted ~\cite{Tuytelaars09}, i.e., Purity and Normalized Mutual Information (NMI). We take these two criteria as forms of empirical similarity examination between two clustering results from adjacent iterations. If the similarity is above a certain threshold, we believe the optimal clustering-based categorization of the data is reached. We indeed find that the final number of categories from the RIM process in later LDPO iterations stabilize around a constant number. The convergence on classification is directly observable through the increasing top-1, top-5 classification accuracy levels in the initial few LDPO rounds which eventually fluctuate slightly at higher values. 

%
%


Convergence in clustering is achieved by adopting the underlying classification capability stored in those deep CNN features through the looped optimization, which accents the visual coherence amongst images inside each cluster. Nevertheless, the category discovery of medical images will further entail clinically semantic labeling of the images. From the optimized clusters, we collect the associated text reports for each image and assemble each cluster's text reports together as a unit. Then NLP is performed on each report unit to find highly recurring words to serve as key word labels for each cluster by simply counting and ranking the frequency of each word. Common words to all clusters are removed from the list. The resultant key words and randomly sampled exemplary images are ultimately compiled for review by board-certified radiologists. This process shares some analogy to the human-machine collaborated image database construction \cite{Yu2015Construction,Wigness2015}. In future work, NLP parsing (especially term negation/assertion) and clustering can be integrated into LDPO framework. 

\subsection{Hierarchical Category Relationship}\label{sec:hcl}

ImageNet \cite{deng2009imagenet} are constructed according to WordNet ontology hierarchy \cite{Miller1995}. Recently, a new formalism so-called Hierarchy and Exclusion (HEX) graphs has been introduced \cite{Deng2014Large} to perform object classification by exploiting the rich structure of real world labels \cite{deng2009imagenet,krizhevsky2012imagenet}. In this work, our converged CNN classification model can be further extended to explore the hierarchical class relationship in a tree representation. First, the pairwise class similarity or affinity score $A_{i,j}$ between class (i,j) is modeled via an adapted measurement from CNN classification confusion \cite{Chen2015Webly}. 
\begin{flalign}
A_{i,j} &= \frac{1}{2} \Big(Prob(i|j) + Prob(j|i) \Big) &\\
&= \frac{1}{2} \Big(\frac{\sum_{I_m\in C_i}CNN(I_m|j)}{|C_i|} + \frac{\sum_{I_n\in C_j}CNN(I_n|i)}{|C_j|}\Big)
\label{eq:HCL:rij}
\end{flalign}
where $C_i$, $C_j$ are the image sets for class $i$,$j$ respectively, $|\cdot|$ is the cardinality function, $CNN(I_m|j)$ is the CNN classification score of image $I_m$ from class $C_i$ at class $j$ obtained directly by the N-way CNN flat-softmax. Here $A_{i,j} = A_{j,i}$ is symmetric by averaging $Prob(i|j)$ and $Prob(j|i)$.

Affinity Propagation algorithm \cite{frey07affinitypropagation} (AP) is invoked to perform ``tuning parameter-free'' clustering on this pairwise affinity matrix $\{A_{i,j}\} \in \mathbb{R}^{K\times K}$. This process can be executed recursively to generate a hierarchically merged category tree. Without loss of generality, we assume that at level L, classes $i^L$,$j^L$ are formed by merging classes at level L-1 through AP clustering. The new affinity score $A_{i^L,j^L}$ is computed as follows.
{\small
	\begin{eqnarray}
	A_{i^L,j^L} = \frac{1}{2} \Big(Prob(i^L|j^L) + Prob(j^L|i^L) \Big)
	\\  Prob(i^L|j^L) = \frac{\sum_{I_m\in C_{i^L}}\sum_{k \in {j^L}}CNN(I_m|k)}{|C_{i^L}|}
	\label{eq:HCL:rijH}
	\end{eqnarray}
}
where L-th level class label ${j^L}$ include all merged original classes (i.e., 0-th level before AP is called) $k \in {j^L}$ so far. From the above, the N-way CNN classification scores (Sec. \ref{sec:cl}) only need to be evaluated once. $A_{i^L,j^L}$ at any level can be computed by summing over these original scores. The discovered category hierarchy can help alleviate the highly uneven visual separability between different object categories in image classification \cite{Yan2015hd} from which the category-embedded hierarchical deep CNN could be beneficial. 

\section{Experimental Results \& Discussion}\label{sec-Exp}

\subsubsection{Dataset:}
We experiment on the same dataset used in~\cite{Shin2015}. 
The image database contains totally ~216K $2D$ key-images which are associated with $\sim62$K unique patients' radiology reports. Key-images are directly extracted from the Dicom file and resized as $256\times256$ bitmap images. Their intensity ranges are rescaled using the default window settings stored in the Dicom header files (this intensity rescaling factor improves the CNN classification accuracies by $\sim2\%$ to ~\cite{Shin2015}). Linked radiology reports are also collected as separate text files with patient-sensitive information removed for privacy reasons. At each LDPO iteration, the image clustering is first applied on the entire image dataset so that each image will receive a cluster label. Then the whole dataset is randomly reshuffled into three subgroups for CNN fine-tuning via Stochastic Gradient Descent (SGD): i.e. training ($70\%$), validation ($10\%$) and testing ($20\%$). In this way, the convergence is not only achieved on a particular data-split configuration but generalized to the entire database.

In order to quantitatively validate our proposed LDPO framework, we also apply category discovery on the texture-25 dataset~\cite{Dai2015EnProDeepFets,Lazebnik2005Sparse}: 25 texture classes, with 40 samples per class. The images from Texture-25 appear drastically different from those natural images in ImageNet, similar to our domain adaptation task from natural to radiology images. The ground truth labels are first hidden from the unsupervised LDPO learning procedure and then revealed to produce the quantitative measures (where purity becomes accuracy) against the resulted clusters. The cluster number is assumed to be known to LDPO and thus the model selection module of RIM in clustering is dropped. 

\subsubsection{CNN Fine-tuning:}
The Caffe~\cite{jia2014caffe} implementation of CNN models are used in the experiment. During the looped optimization process, the CNN is fine-tuned for each iteration once a new set of image labels is generated from the clustering stage. Only the last softmax classification layer of the models (i.e. 'FC8' in AlexNet and 'loss3/classifier' in GoogLeNet) is significantly modulated by 1) setting a higher learning rate than all other layers and 2) updating the (varying but converging) number of category classes from the newly computed results of clustering. 

\subsection{LDPO Convergence Analysis} 
We first study how the different settings of proposed LDPO framework will affect  convergence as follows:  

\subsubsection{Clustering Method:}
\begin{figure*}[t]
	\begin{center}
		\includegraphics[width=0.325\linewidth]{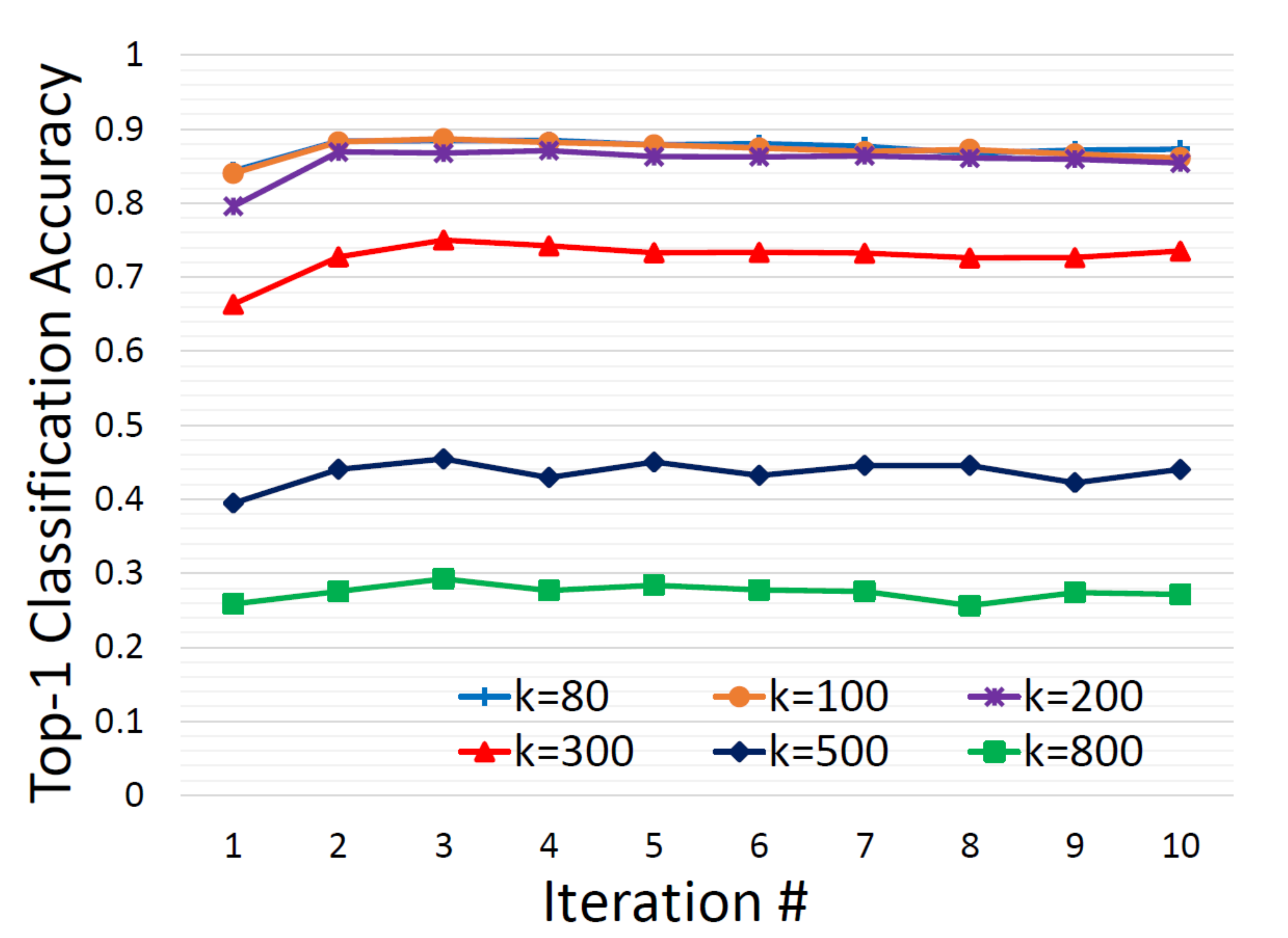}
		\includegraphics[width=0.325\linewidth]{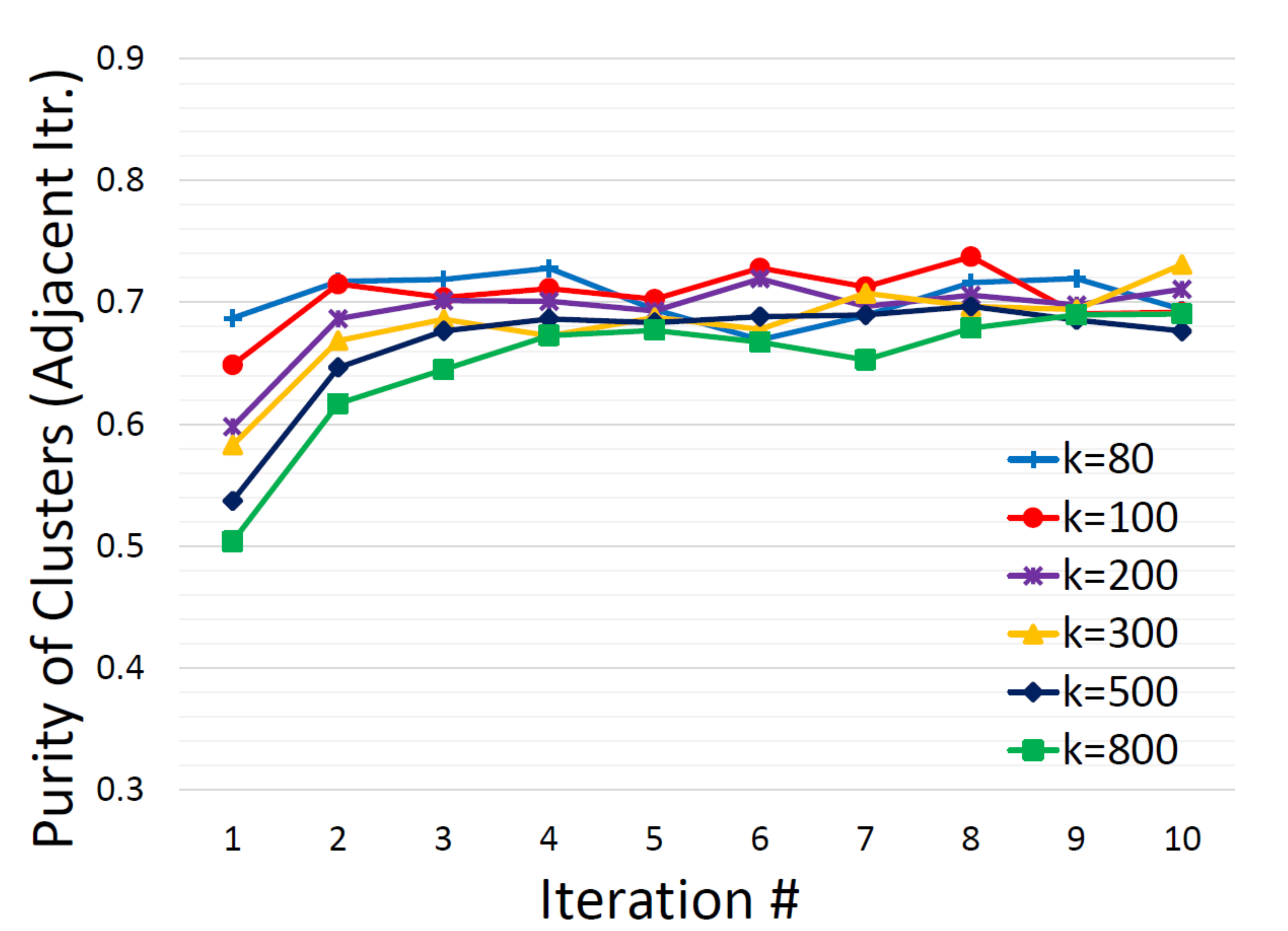}
		\includegraphics[width=0.325\linewidth]{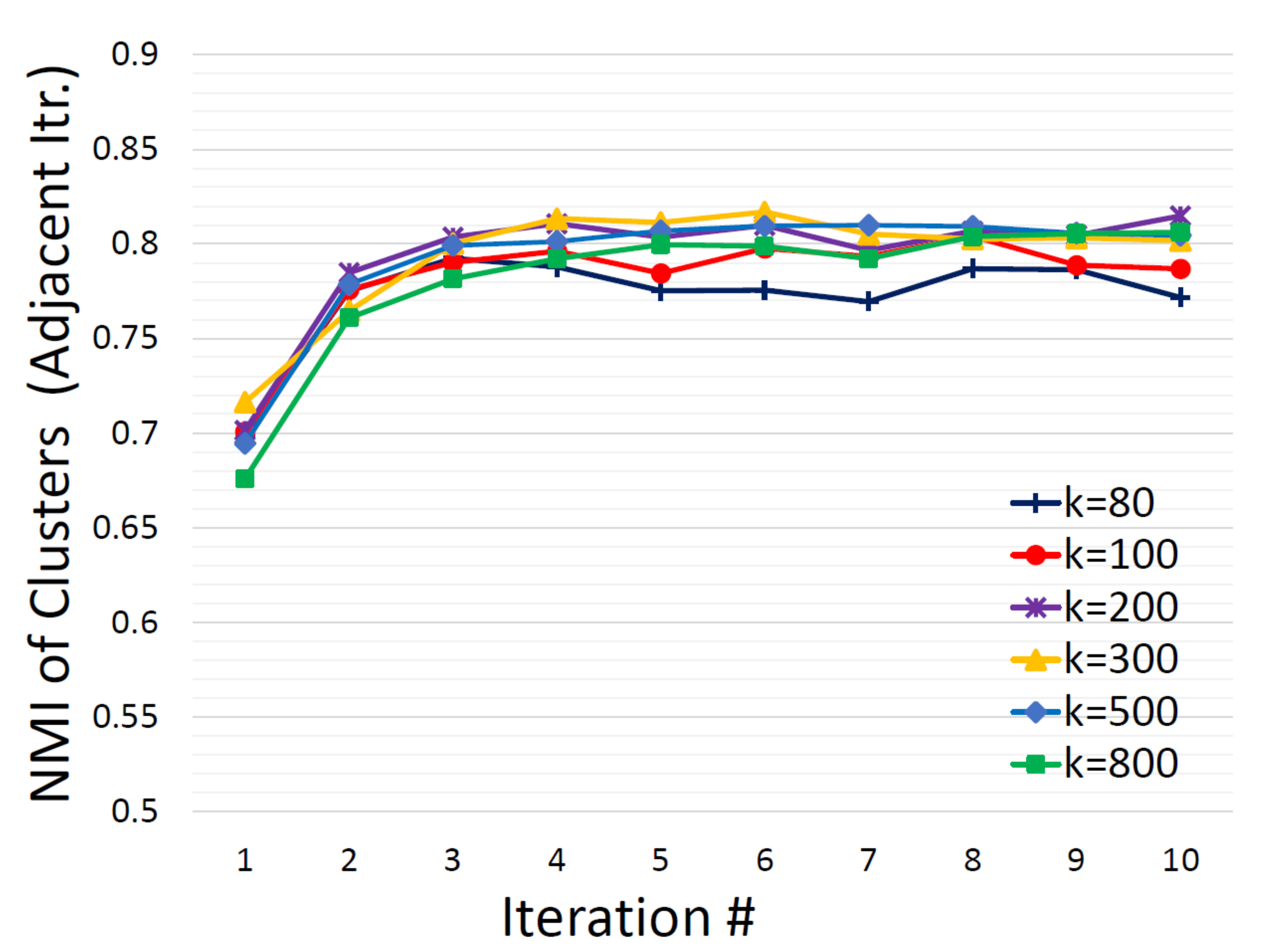}		
	\end{center}
	\caption{Performance of LDPO using $k$-means clustering with a variety of $k$. From left to right, the top-1 classification accuracy and the purity and NMI of clusters from adjacent iterations are shown.} 
	\label{fig:kmeans:png}
\end{figure*}

\begin{figure*}[t]
	\begin{center}
		\includegraphics[width=0.47\linewidth]{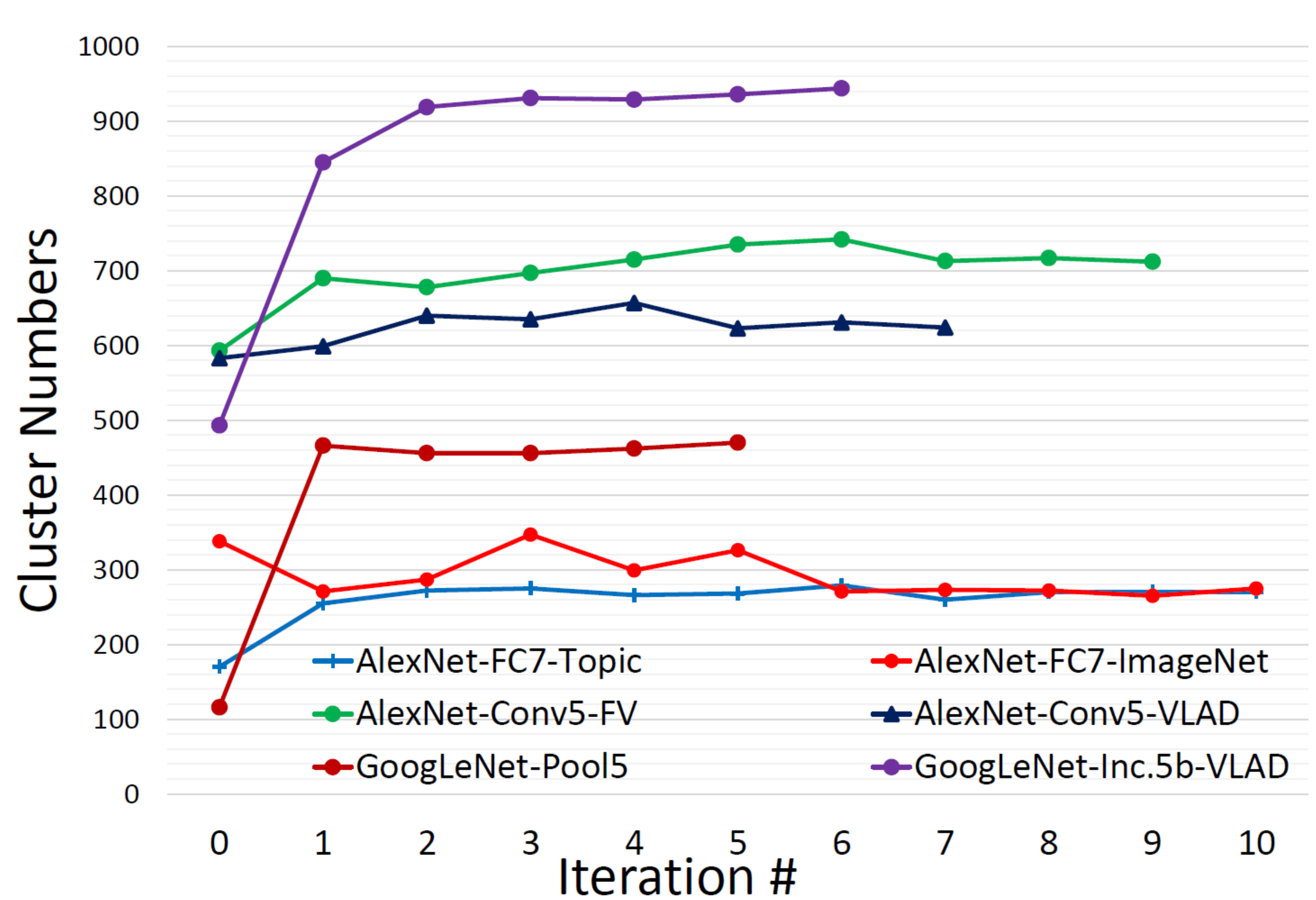}
		\includegraphics[width=0.47\linewidth]{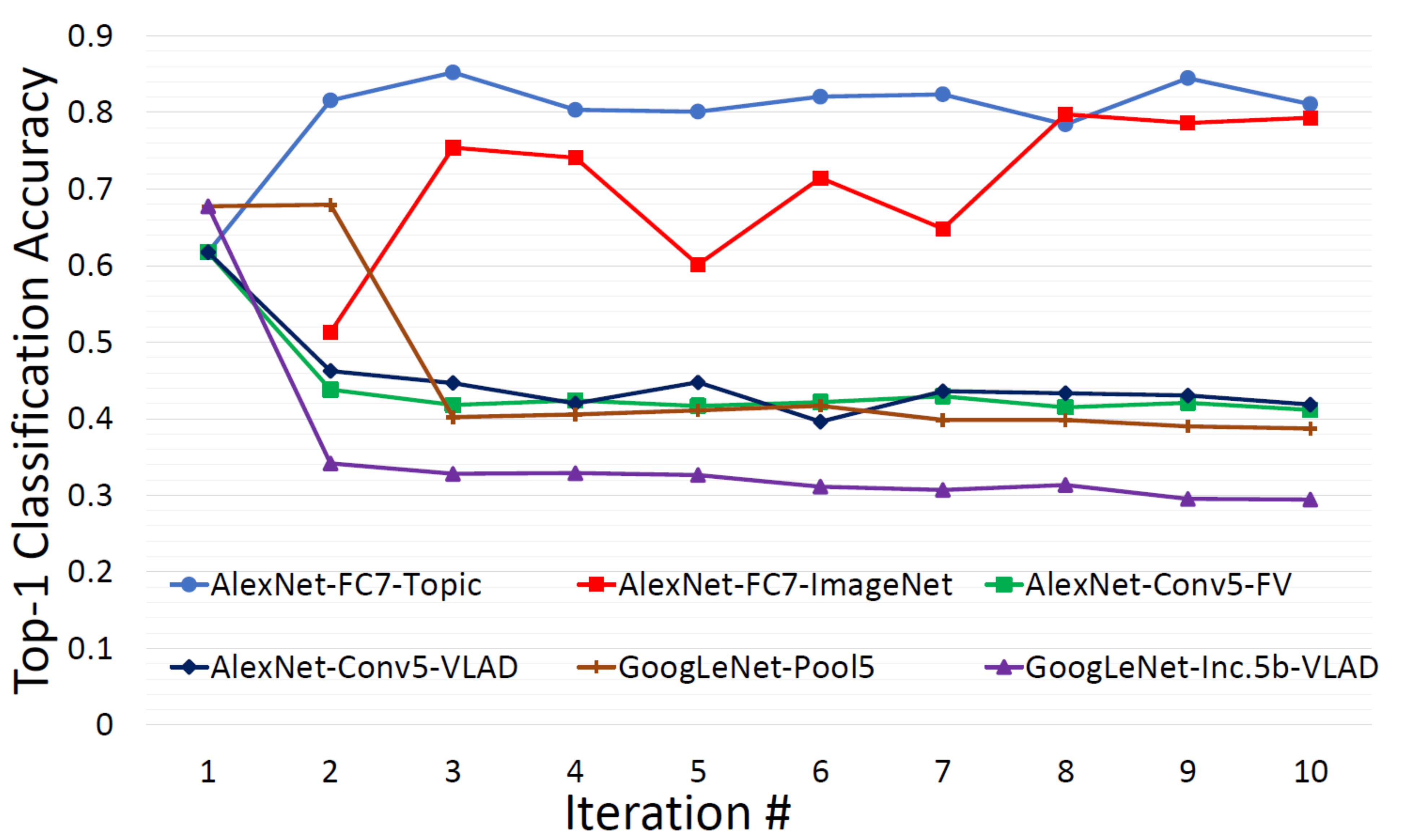}\\
		\includegraphics[width=0.47\linewidth]{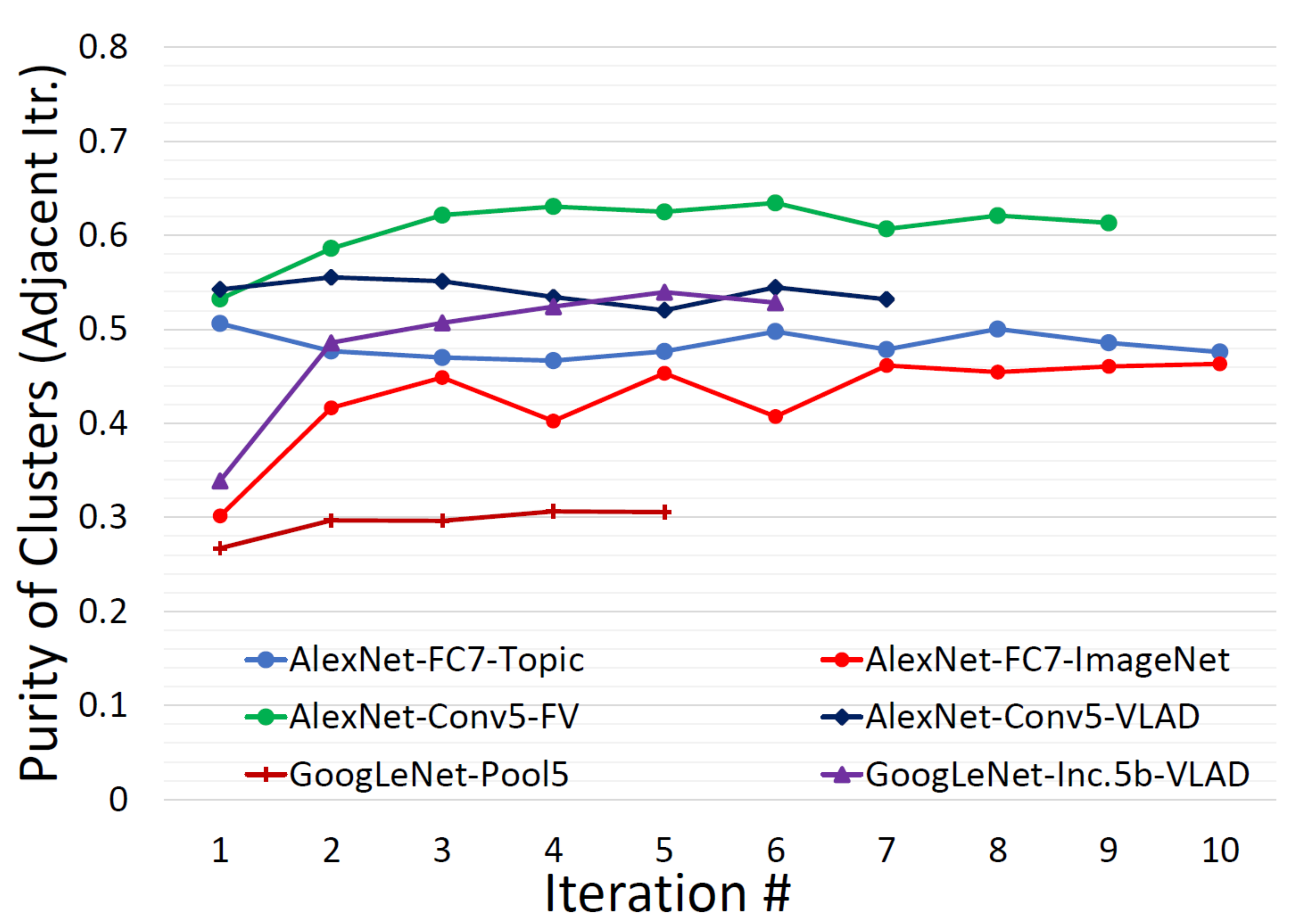}
		\includegraphics[width=0.47\linewidth]{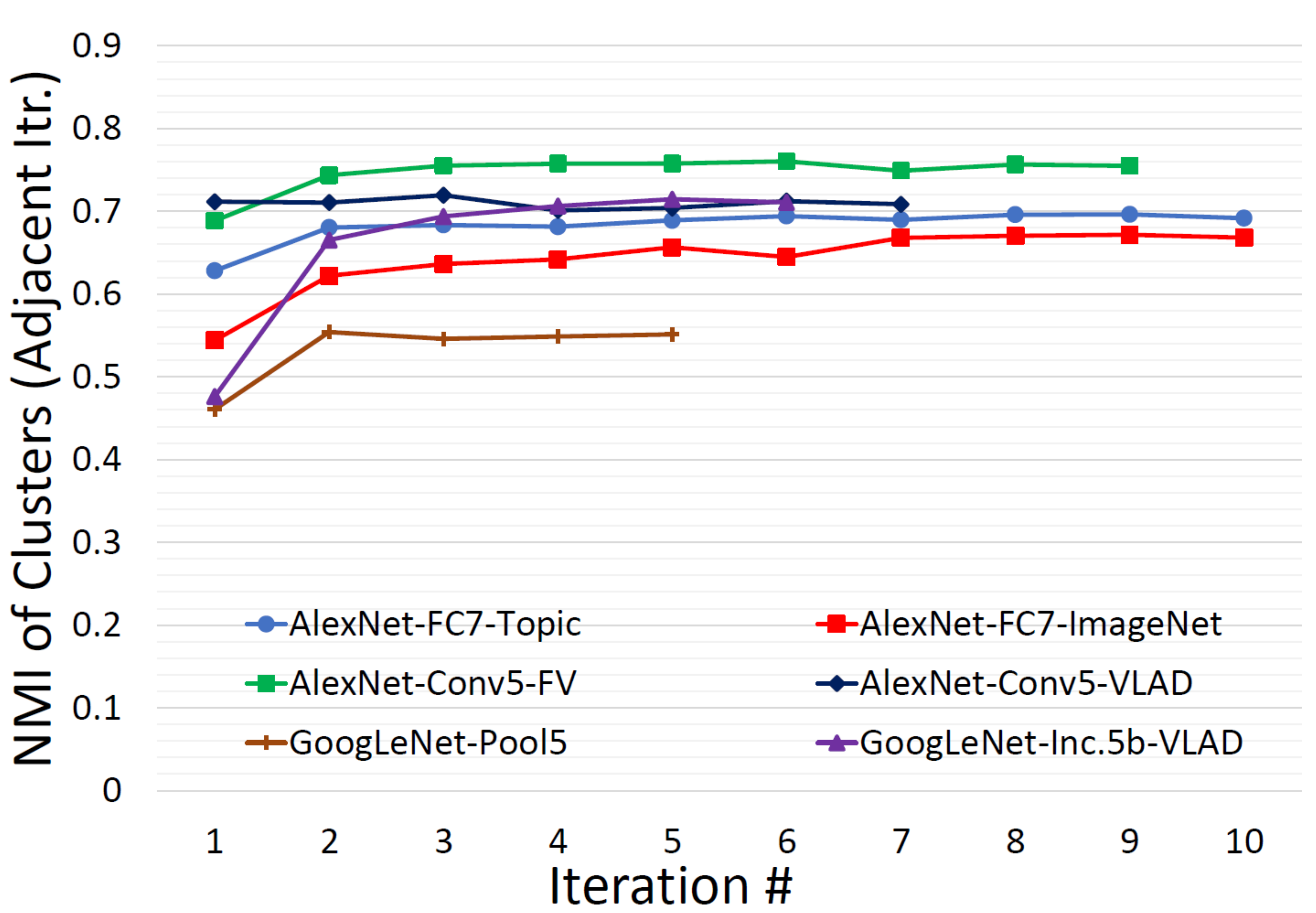}		
	\end{center}
	\caption{Performance of LDPO using RIM clustering with different image encoding methods (\textit{i.e.}, FV and VLAD) and CNN Architectures (\textit{i.e.}, AlexNet and GoogLeNet). From left to right(top to bottom), the number of clusters discovered, Top-1 accuracy of trained CNNs, the purity and NMI of clusters from adjacent iterations are illustrated.}
	\label{fig:RIM_Encode:png}
\end{figure*}

We perform $k$-means based image clustering with $k \in$  $\{80,100,$  $200,300,500,800\}$. Fig.~\ref{fig:kmeans:png} shows the changes of top-1 accuracy, cluster purity and NMI with different $k$ across iterations. The classification accuracies quickly plateau after 2 or 3 iterations. Smaller $k$ values naturally trigger higher accuracies ($>86.0$\% for $k=80$) as less categories make the classification task easier. Levels of Purity and NMI between clusters from two consecutive iterations increase quickly and fluctuate close to $0.7$, thus indicating the convergence of clustering labels (and CNN models). The minor fluctuation are rather due to the randomly re-sorting of the dataset in each iteration. RIM clustering takes an over-segmented $k$-means results as initialization, e.g., $k=1000$ in our experiments. As shown in Fig.~\ref{fig:RIM_Encode:png} Top-left, RIM can estimate the category capacities or numbers consistently under different image representations (deep CNN feature + encoding approaches). $k$-means clustering enables LDPO to approach the convergence quickly with high classification accuracies; whereas, the added RIM based model selection delivers more balanced and semantically meaningful clustering results (see more in Sec.~\ref{sec-label-results}). This is due to RIM's two unique characteristics: 1), less restricted geometric assumptions in the clustering feature space; 
2), the capacity to attain the optimal number of clusters by maximizing the mutual information of input data and the induced clusters via a regularized term. 

\begin{figure*}[!t]
	\begin{center}
		\includegraphics[width=0.31\linewidth]{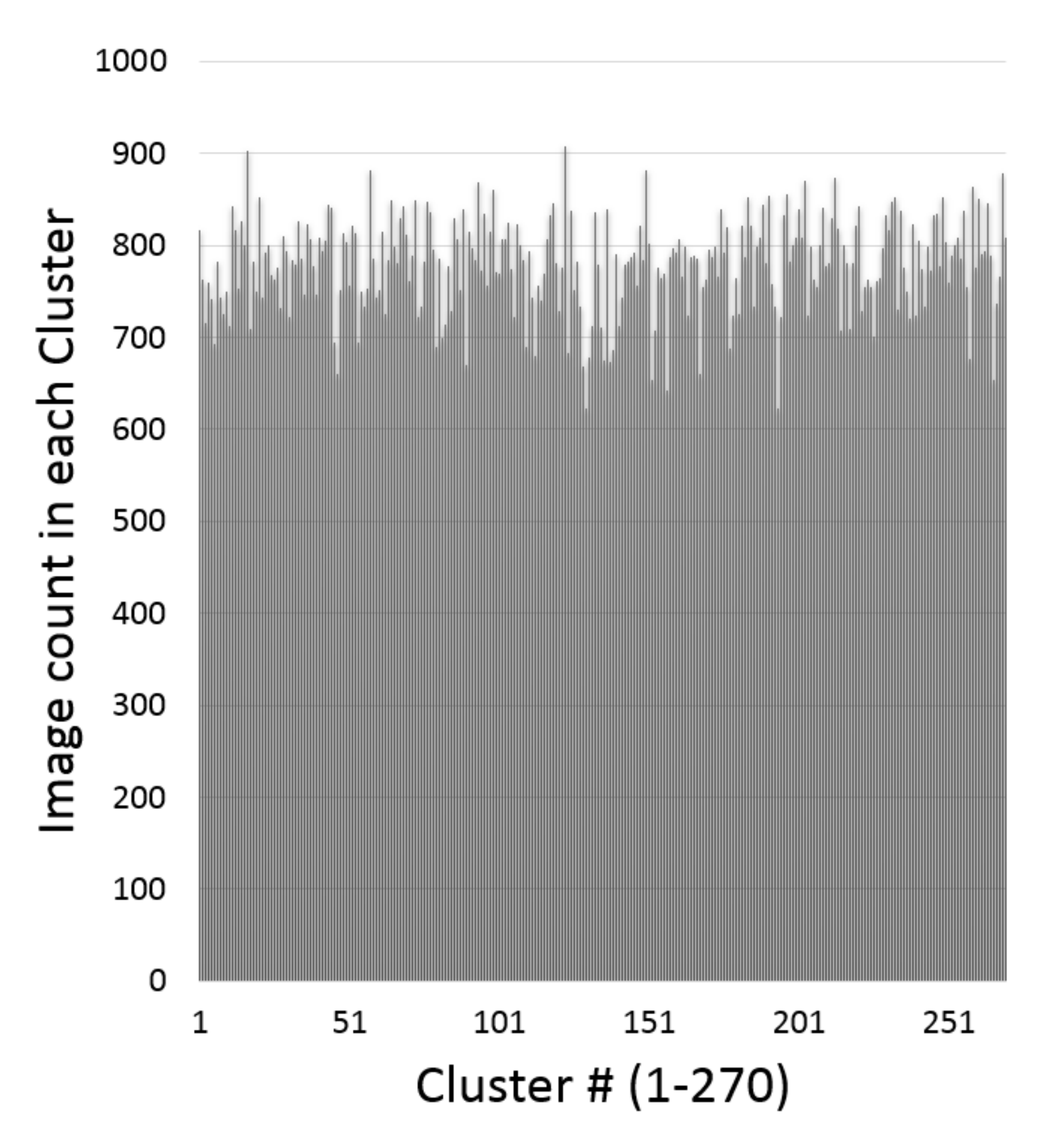}
		\includegraphics[width=0.61\linewidth]{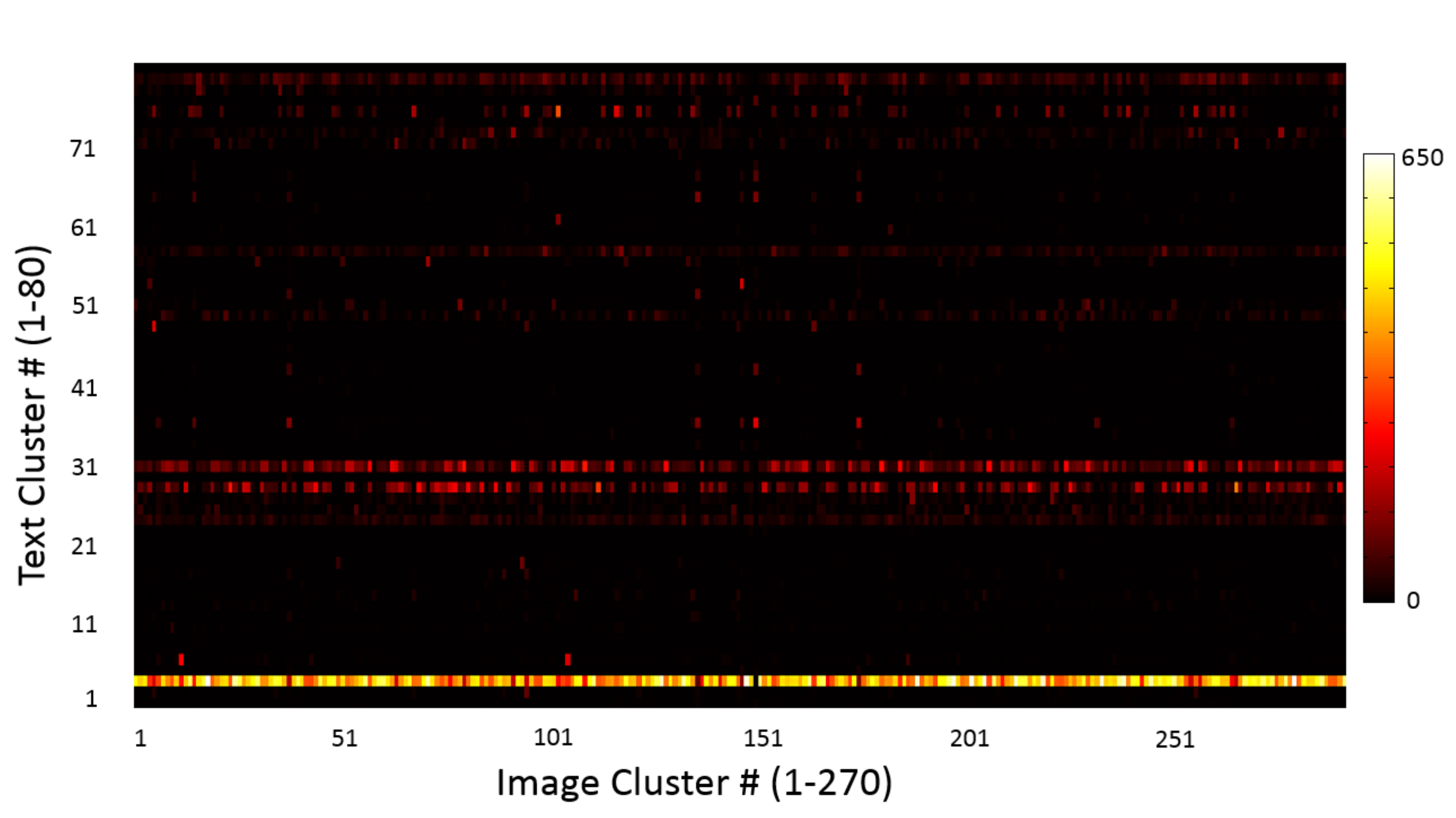}		
	\end{center}
	\caption{Statistics of converged categories using the Alexnet-FC7-Topic setting. Left: the image numbers in each cluster; Right: affinity matrix of two clustering results (AlexNet-FC7-270 vs Text-Topics-80 produced using the approach in~\cite{Shin2015}).} 
	\label{fig:Clusters:png}
\end{figure*}

\subsubsection{Pseudo-Task Initialization:}
Both ImageNet and domain-specific~\cite{Shin2015} CNN models have been employed to initialize the LDPO framework. In Fig.~\ref{fig:RIM_Encode:png}, two CNNs of AlexNet-FC7-ImageNet and AlexNet-FC7-Topic demonstrate their LDPO performances. LDPO initialized by ImageNet CNN reach the steady state noticeably slower than its counterpart, as AlexNet-FC7-Topic already contains the domain information from this radiology image database. However, similar clustering outputs are produced after convergence. Letting LDPO reach $\sim10$ iterations, two different initializations end up with very close clustering results (i.e., Cluster number, purity and NMI) and similar classification accuracies (shown in Table~\ref{tab:CNN-Acc}).  

\subsubsection{CNN Deep Feature and Image Encoding:}
Different image representations can vary the performance of proposed LDPO framework as shown in Fig.~\ref{fig:RIM_Encode:png}. As mentioned in Sec.~\ref{sec-method}, deep CNN images features extracted from different layers of CNN models (AlexNet and GoogLeNet) contain the level-specific visual information. Convolutional layer features retain the spatial activation layouts of images while FC layer features do not. Different encoding approaches further lead to various outcomes of our LDPO framework. The numbers of clusters range from 270 ({\bf AlexNet-FC7-Topic} with no deep feature encoding) to 931 (the more sophisticated {\bf GoogLeNet-Inc.5b-VLAD} with VLAD encoding). The numbers of clusters discovered by RIM reflect the amount of information complexity stored in the radiology database.  

\subsubsection{Computational Cost:}
LDPO runs on a node of Linux computer cluster with 16 CPU cores (x2650), 128G memory and Nvidia K20 GPUs. The Computational costs of different LDPO configurations are shown in Table~\ref{tab:CNN-Acc} per looped iteration. The more sophisticated and feature rich settings, e.g., {\bf AlexNet-Conv5-FV}, {\bf GoogLeNet-Pool5} and {\bf GoogLeNet-Inc.5b-VLAD}, require more time to converge. 
{\small
	\begin{table}[!t]
	\caption{Classification Accuracy of Converged CNN Models}
	\centering
	\begin{tabular}{|l||c|c|c|}
		\hline\hline
		{\bf \small CNN setting} & {\bf Cluster \#} & {\bf Top-1} & {\bf Top-5} \\[0.3ex]
		\hline\hline
		{\bf \small AlexNet-FC7-Topic}    & 270 & 0.8109 & 0.9412  \\[0.3ex]
		\hline
		{\bf \small AlexNet-FC7-ImageNet}    & 275 & 0.8099 & 0.9547  \\[0.3ex]
		\hline
		{\bf \small AlexNet-Conv5-FV}   & 712  & 0.4115 & 0.4789  \\[0.3ex]
		\hline
		{\bf \small AlexNet-Conv5-VLAD}   & 624 & 0.4333 & 0.5232  \\[0.3ex]
		\hline
		{\bf \small GoogLeNet-Pool5}    & 462 & 0.4109 & 0.5609 \\[0.3ex]
		\hline
		{\bf \small GoogLeNet-Inc.5b-VLAD}   & 929 & 0.3265 & 0.4001 \\[0.3ex]
		\hline 
	\end{tabular}\label{tab:CNN-Acc}
\end{table}
}
\begin{table}[!t]
	\caption{Computational Cost of LDPO}
	\centering
	\begin{tabular}{|l||c|}
		\hline\hline
		{\bf \small CNN setting} & {\bf Time per iter.(HH:MM)} \\[0.3ex]
		\hline\hline
		{\bf \small AlexNet-FC7-Topic}    & 14:35  \\[0.3ex]
		\hline
		{\bf \small AlexNet-FC7-Imagenet}   & 14:40  \\[0.3ex]
		\hline
		{\bf \small AlexNet-Conv5-FV}   & 17:40  \\[0.3ex]
		\hline
		{\bf \small AlexNet-Conv5-VLAD}   & 15:44  \\[0.3ex]
		\hline
		{\bf \small GoogLeNet-Pool5}    & 21:12  \\[0.3ex]
		\hline
		{\bf \small GoogLeNet-Inc.5b-VLAD}   & 23:35  \\[0.3ex]
		\hline 
	\end{tabular}\label{tab:CNN-Time}
\end{table}


\begin{figure*}
	\begin{center}
		\includegraphics[width=0.165\linewidth]{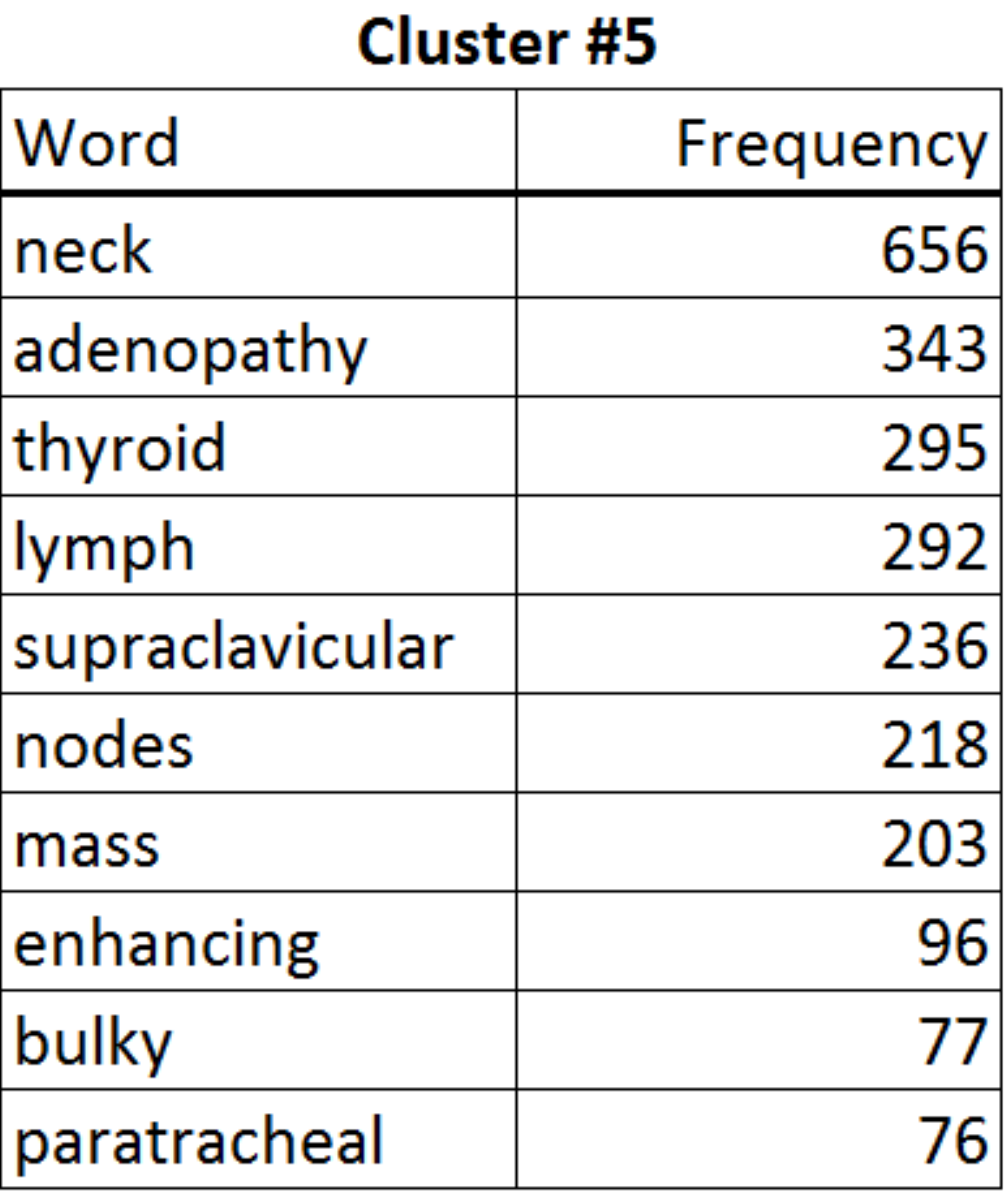}
		\includegraphics[width=0.20\linewidth]{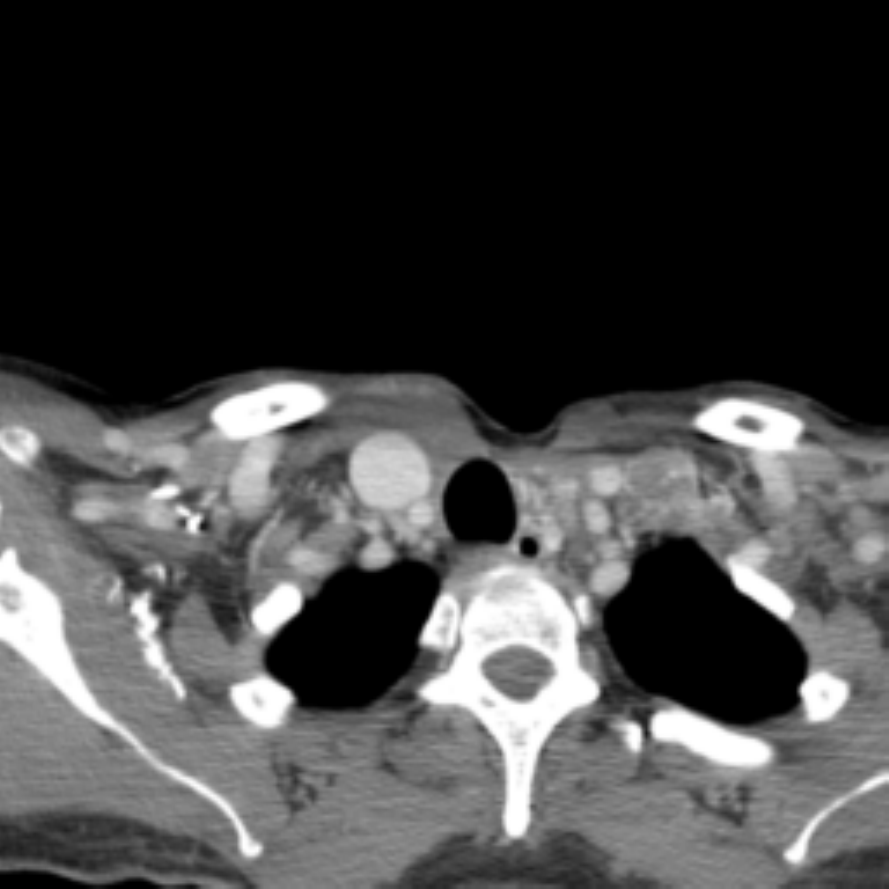}
		\includegraphics[width=0.20\linewidth]{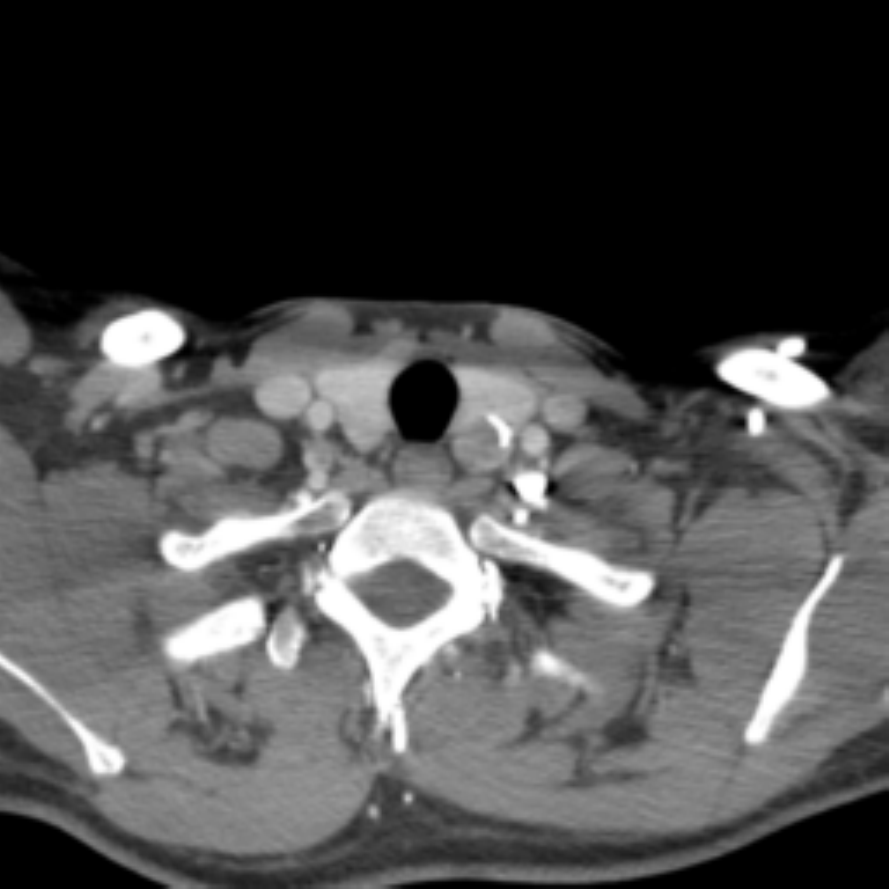}
		\includegraphics[width=0.20\linewidth]{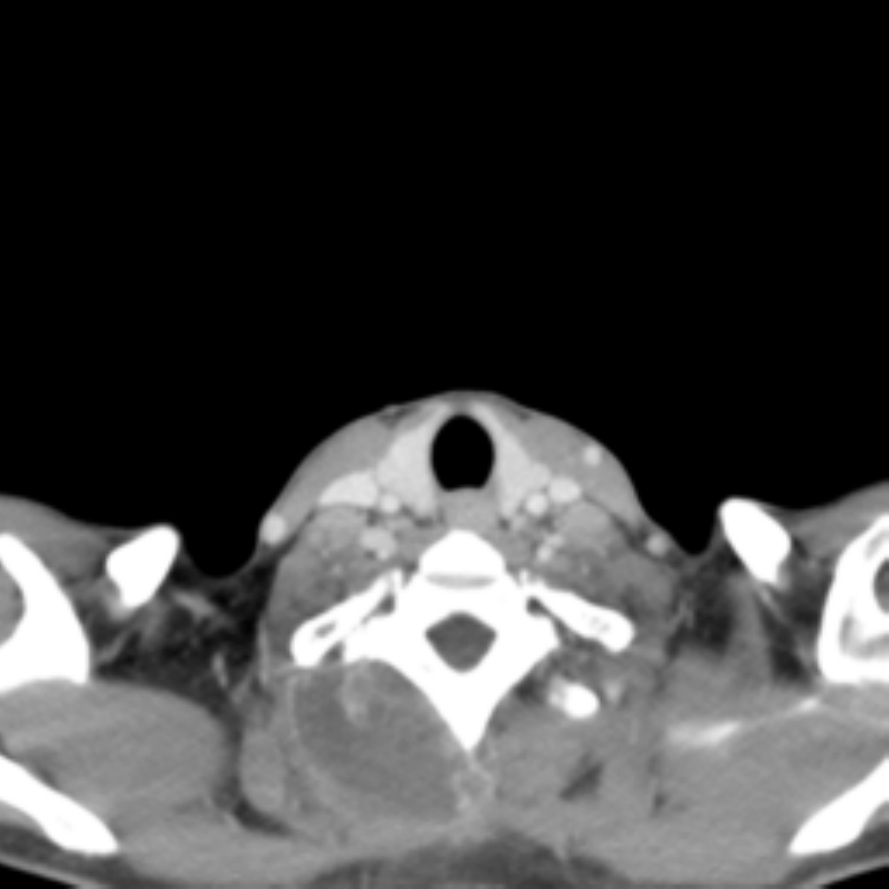}
		\includegraphics[width=0.20\linewidth]{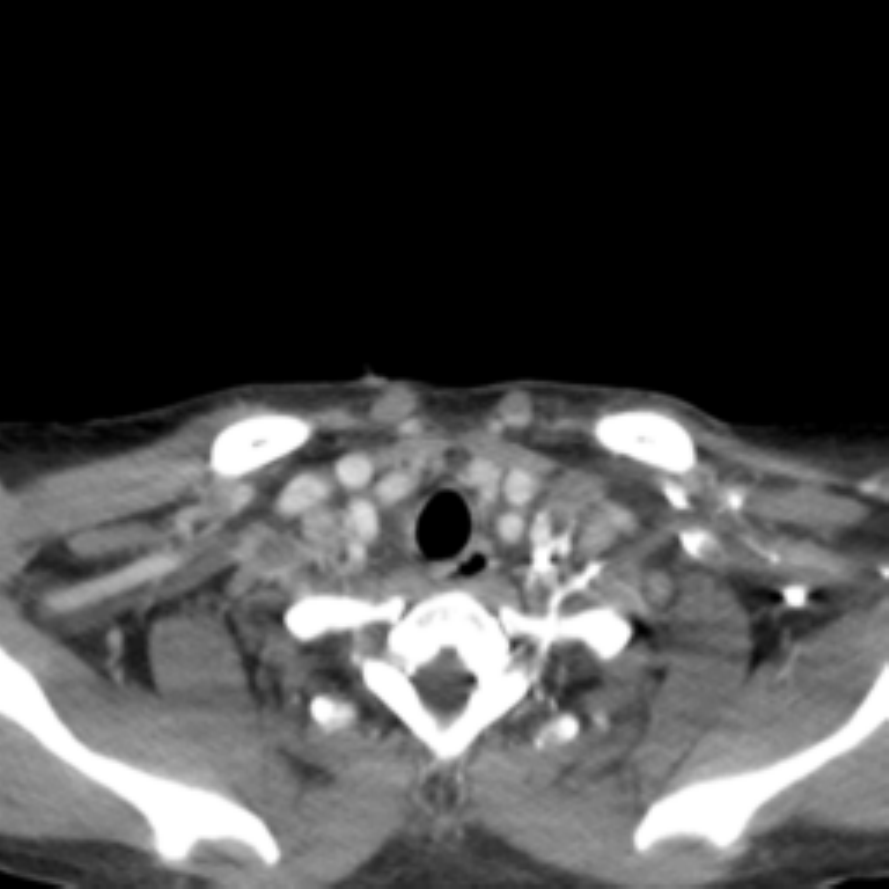}
		\includegraphics[width=0.165\linewidth]{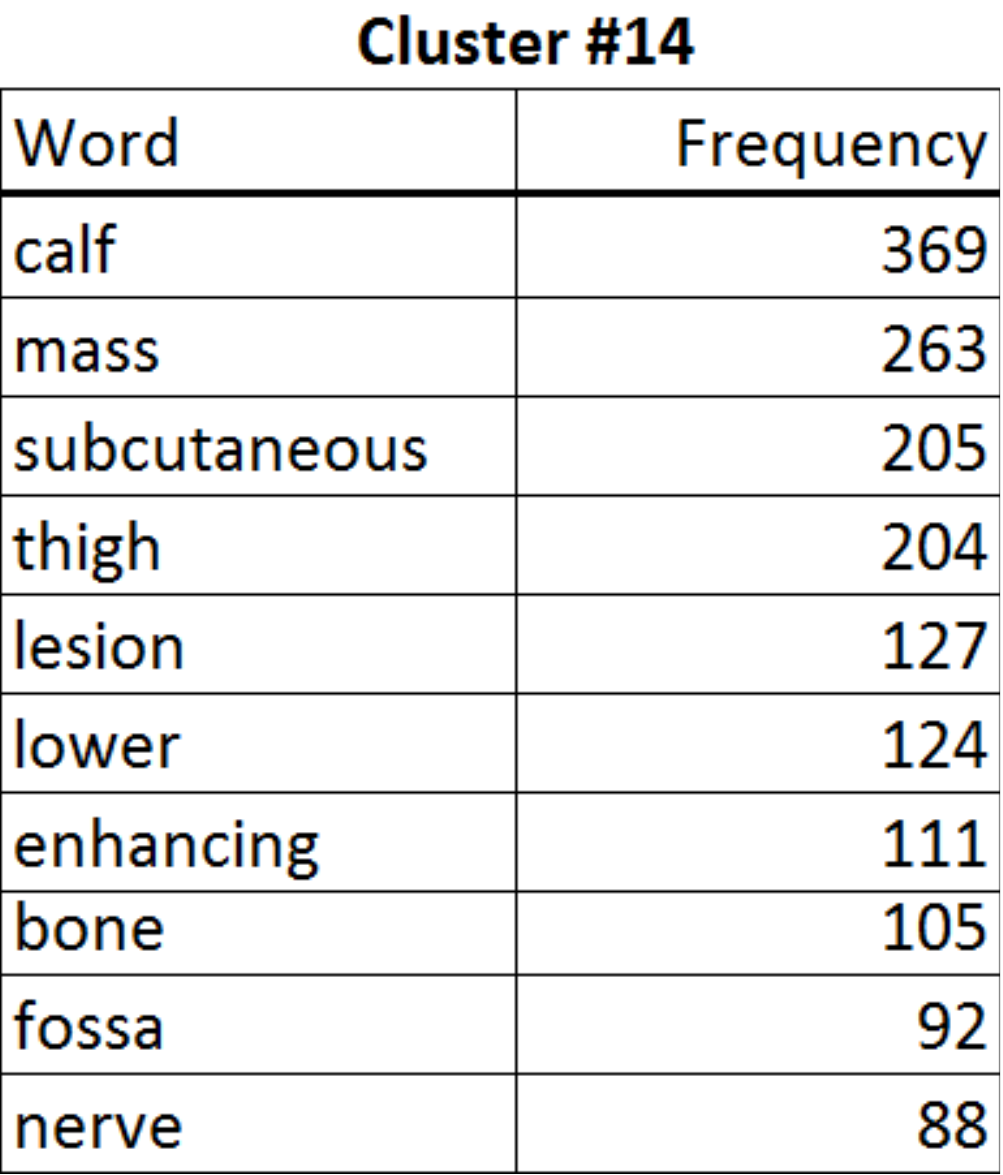}
		\includegraphics[width=0.20\linewidth]{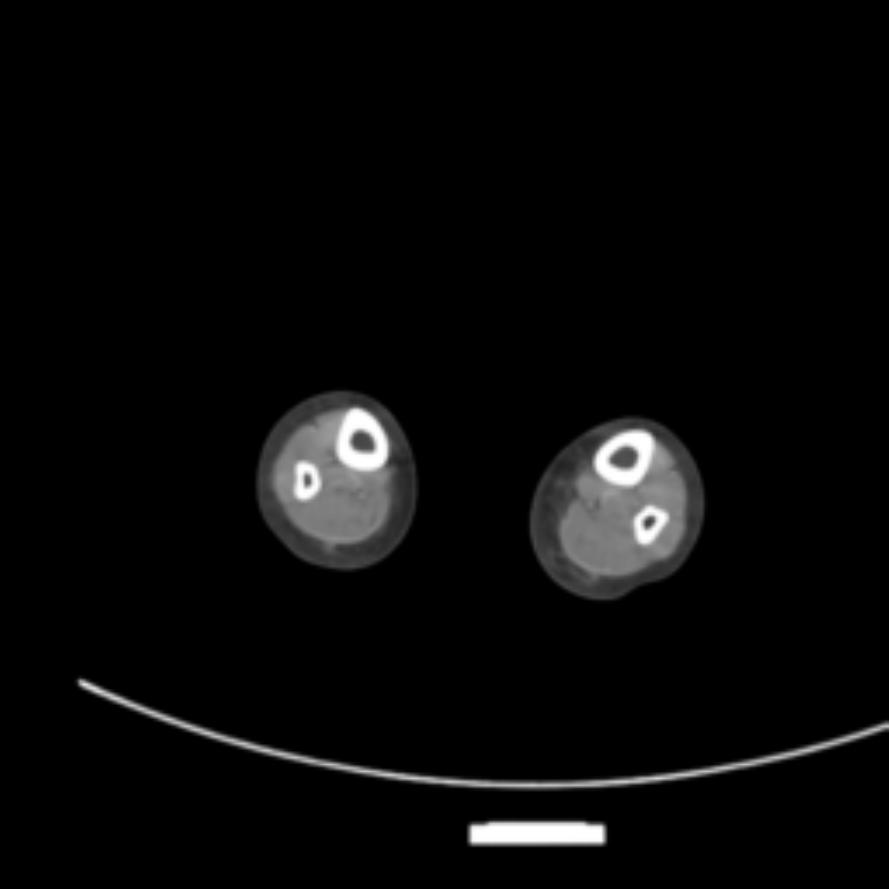}
		\includegraphics[width=0.20\linewidth]{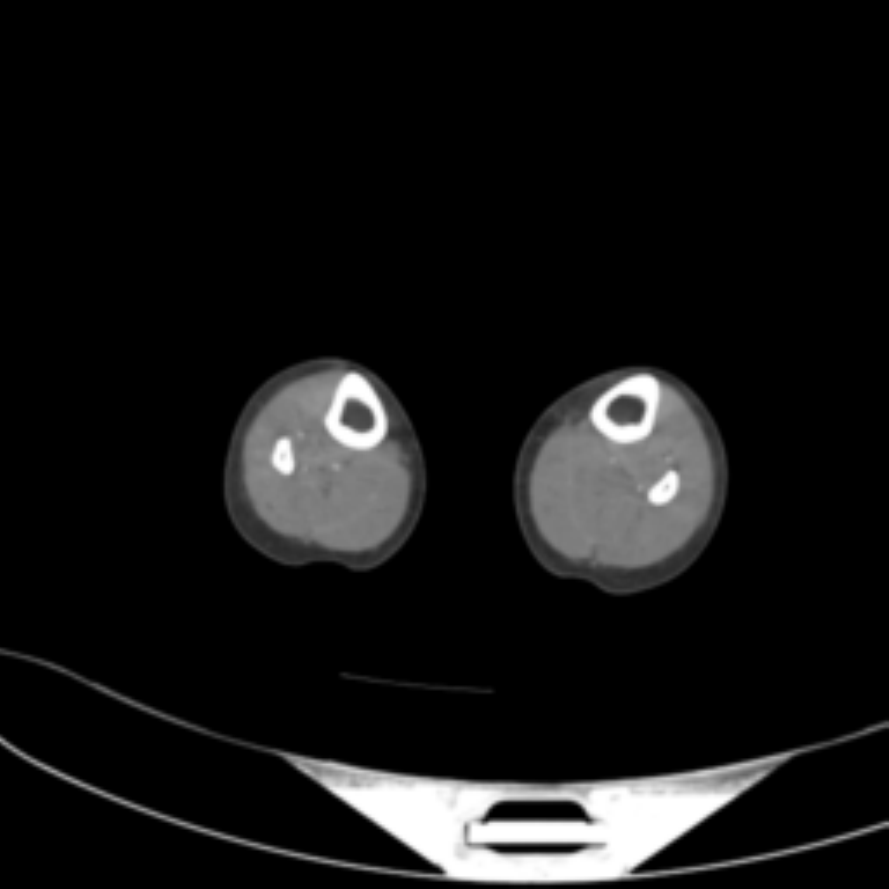}
		\includegraphics[width=0.20\linewidth]{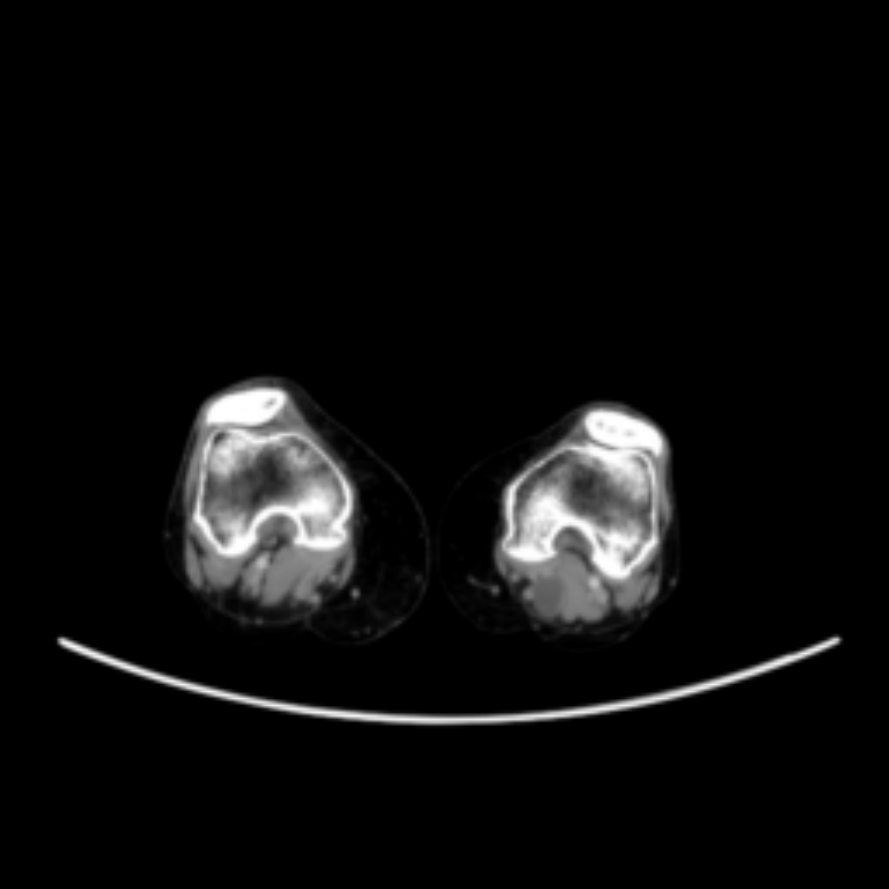}
		\includegraphics[width=0.20\linewidth]{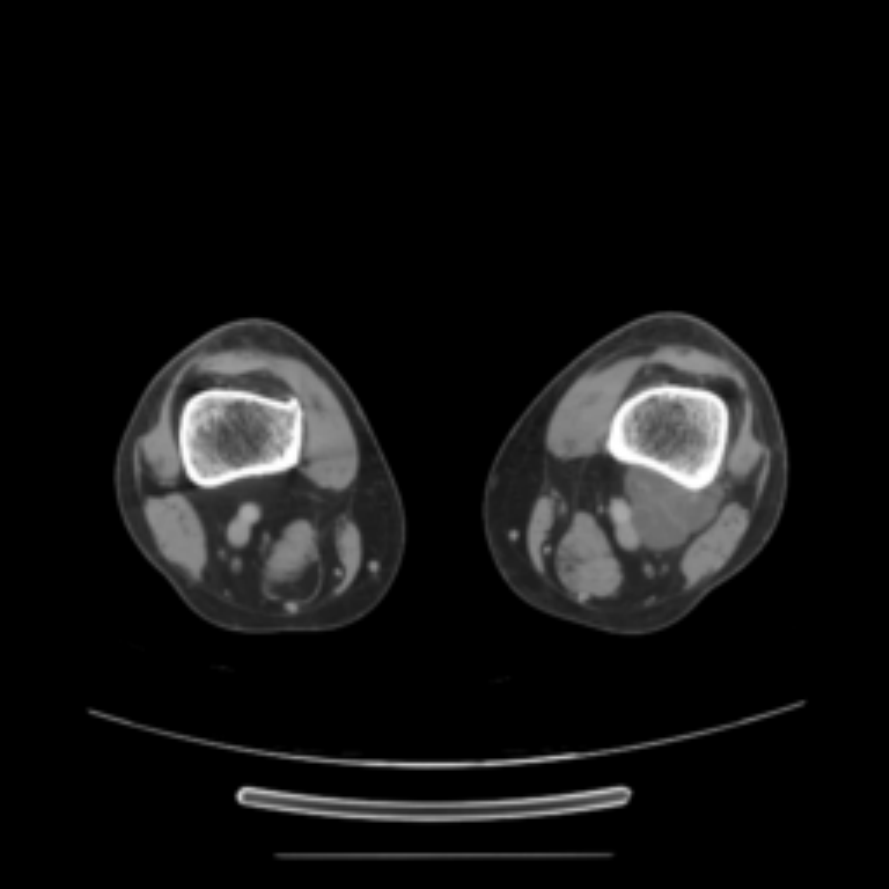}	
		\includegraphics[width=0.165\linewidth]{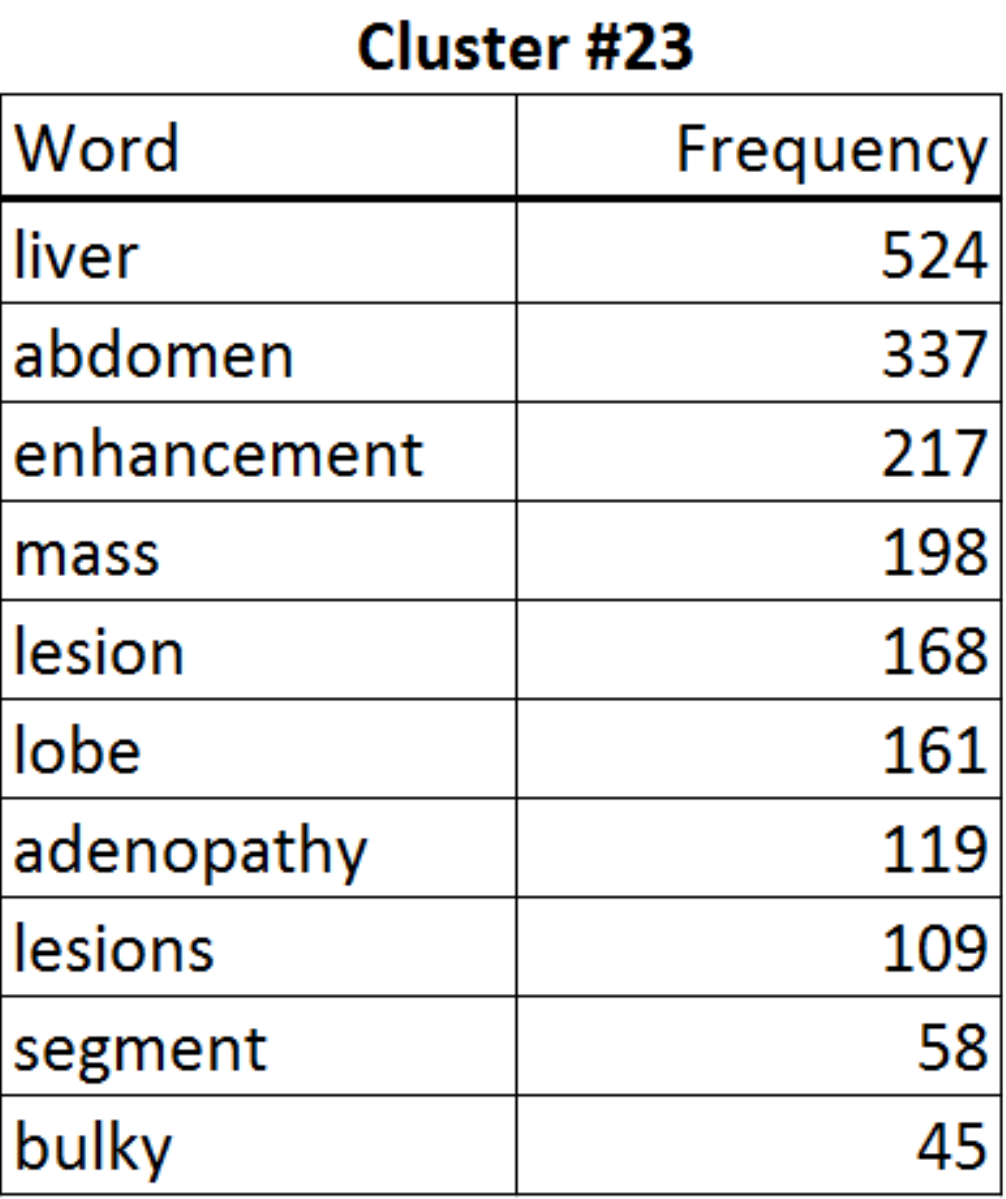}
		\includegraphics[width=0.20\linewidth]{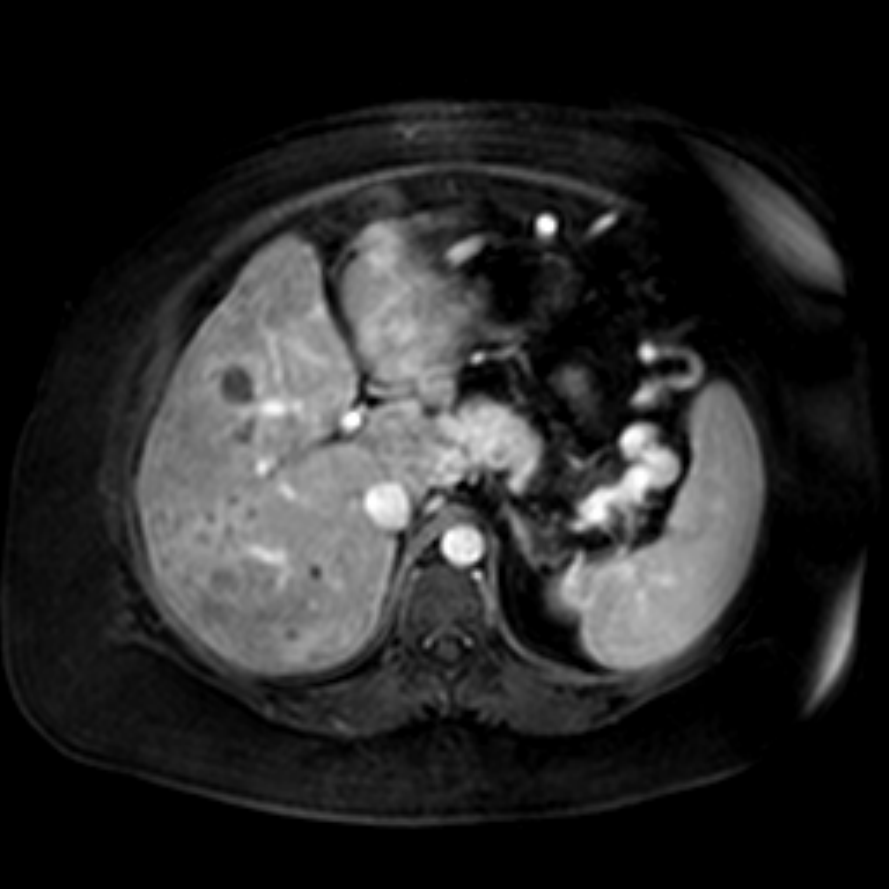}
		\includegraphics[width=0.20\linewidth]{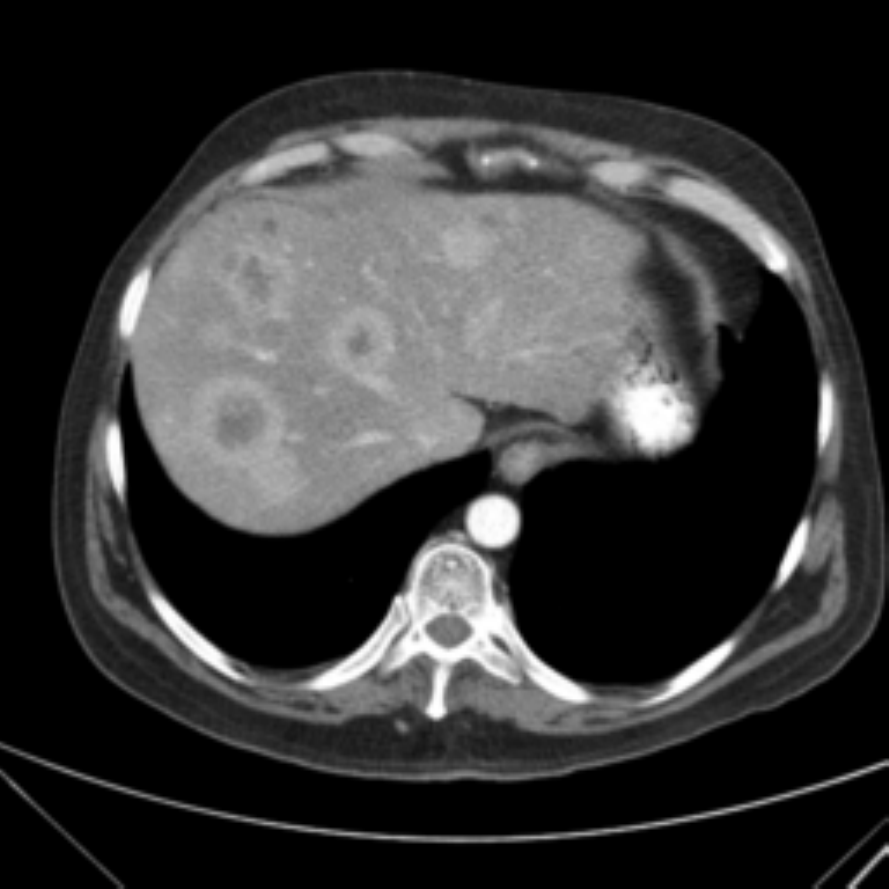}
		\includegraphics[width=0.20\linewidth]{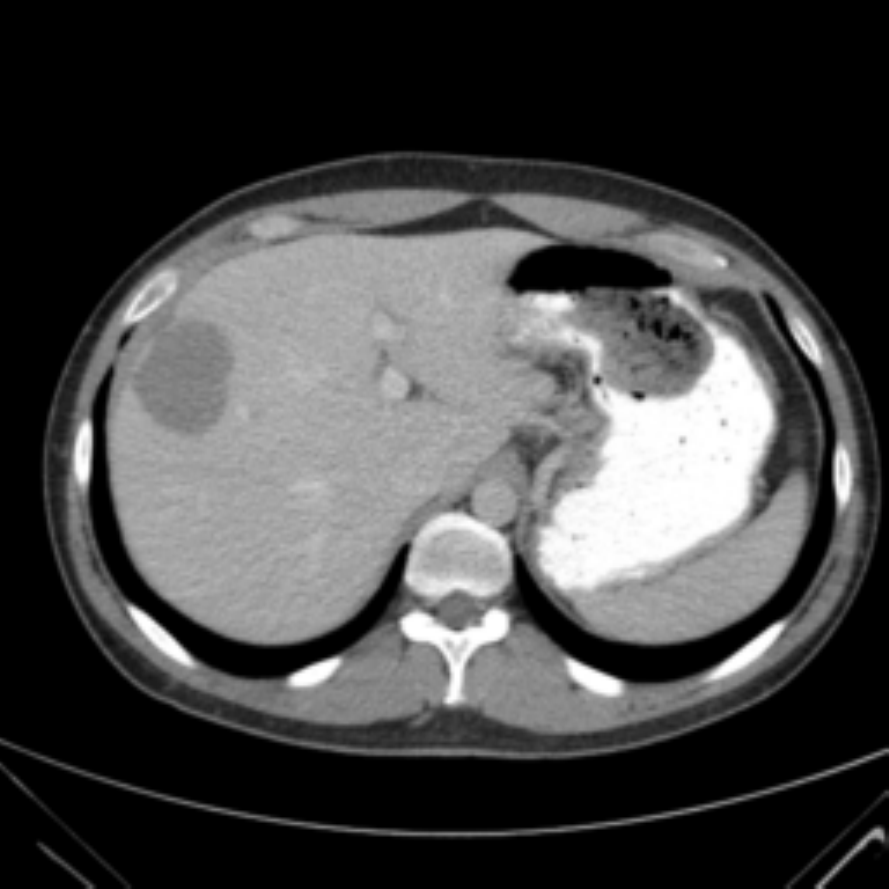}
		\includegraphics[width=0.20\linewidth]{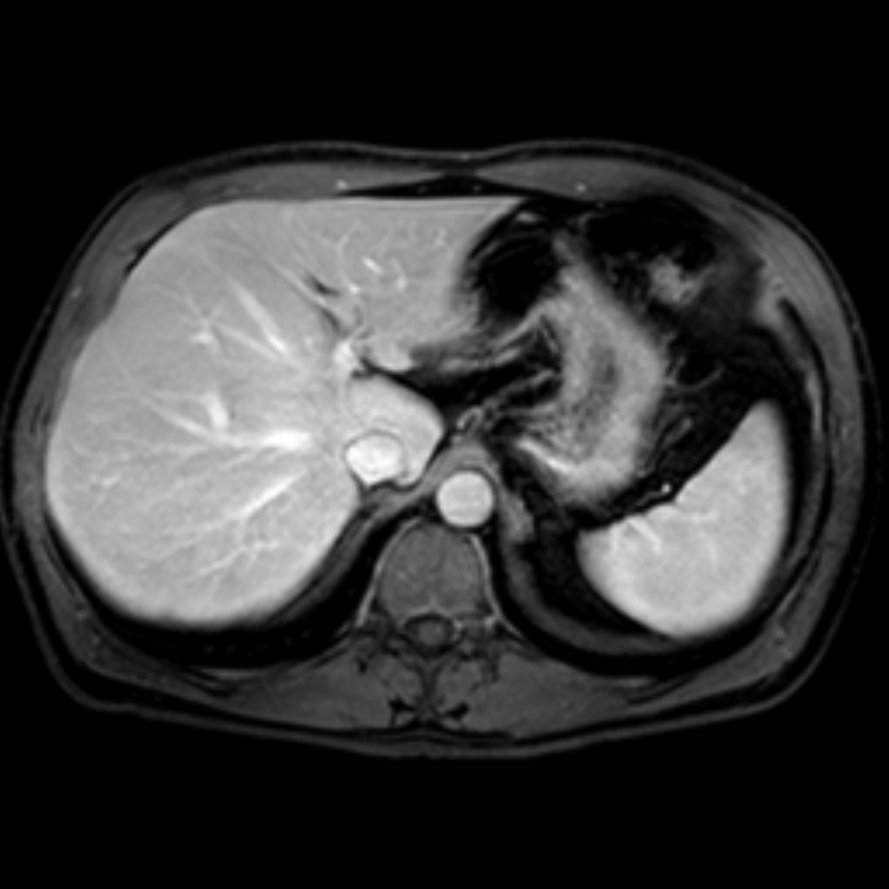}		
		\includegraphics[width=0.165\linewidth]{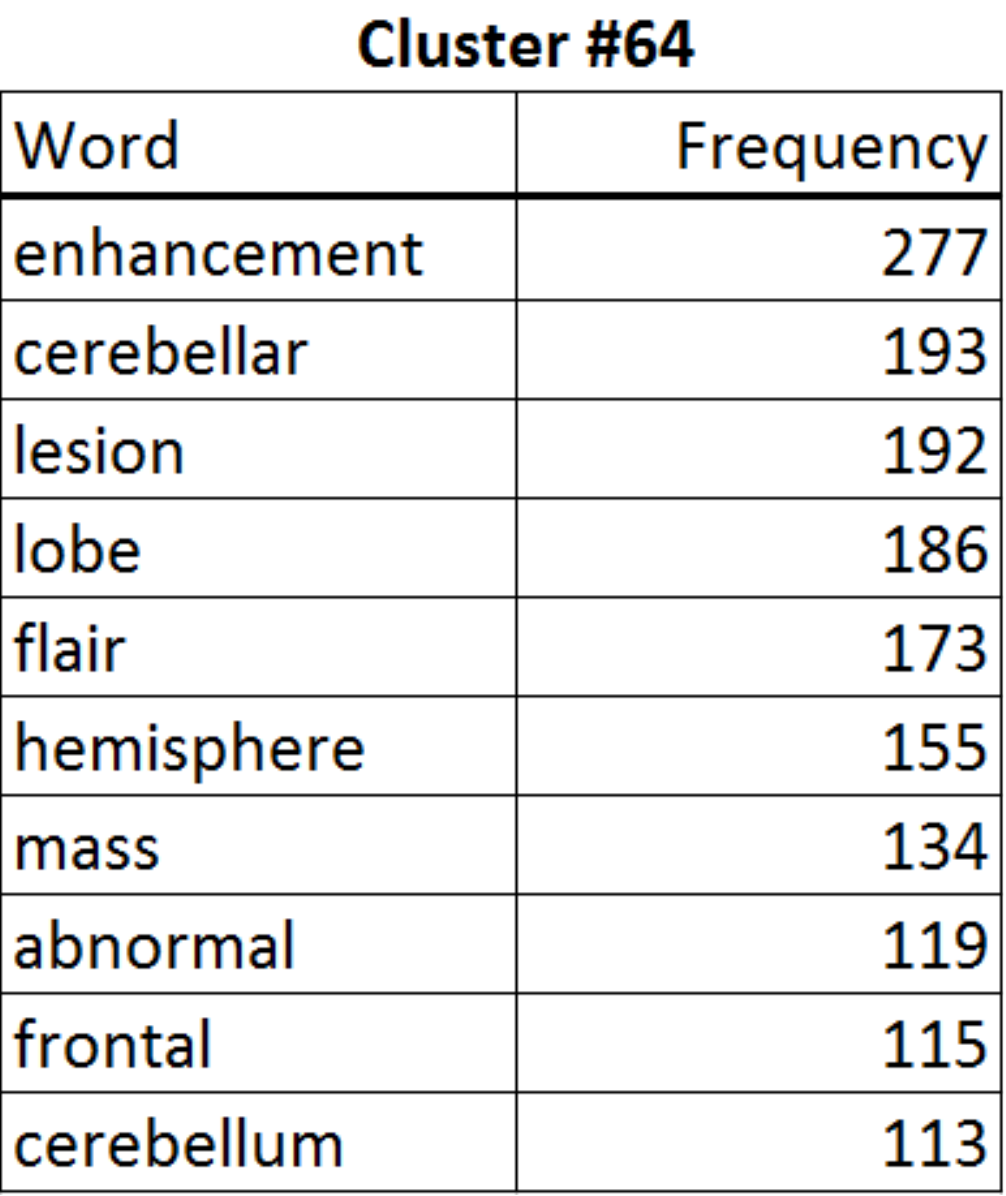}
		\includegraphics[width=0.20\linewidth]{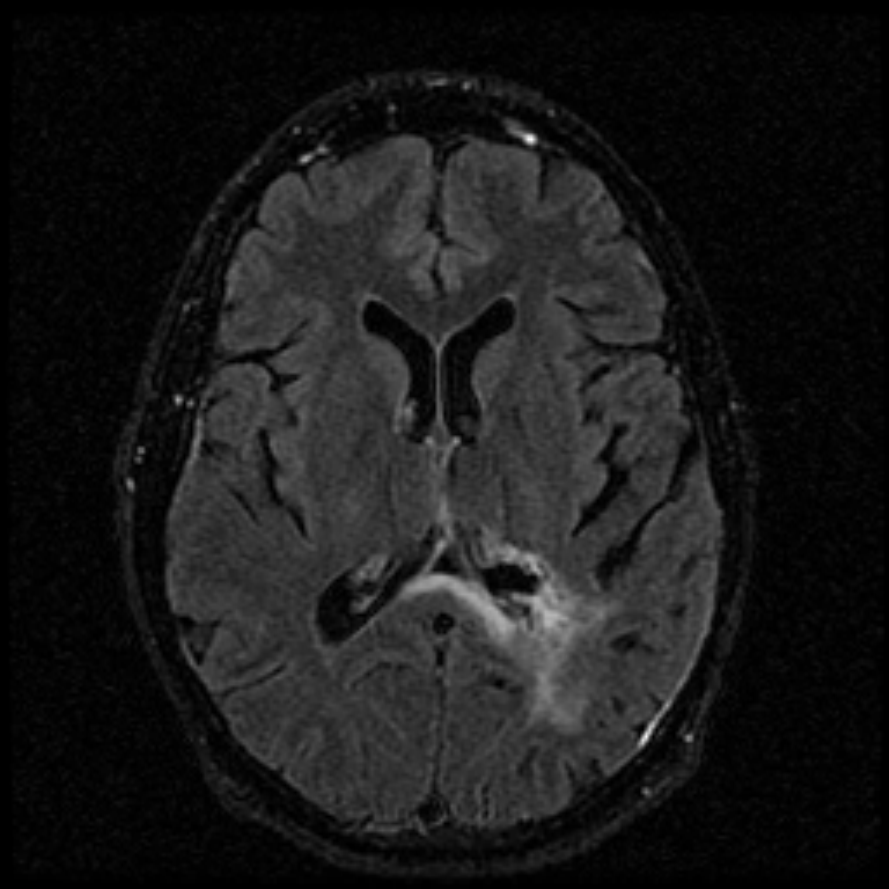}
		\includegraphics[width=0.20\linewidth]{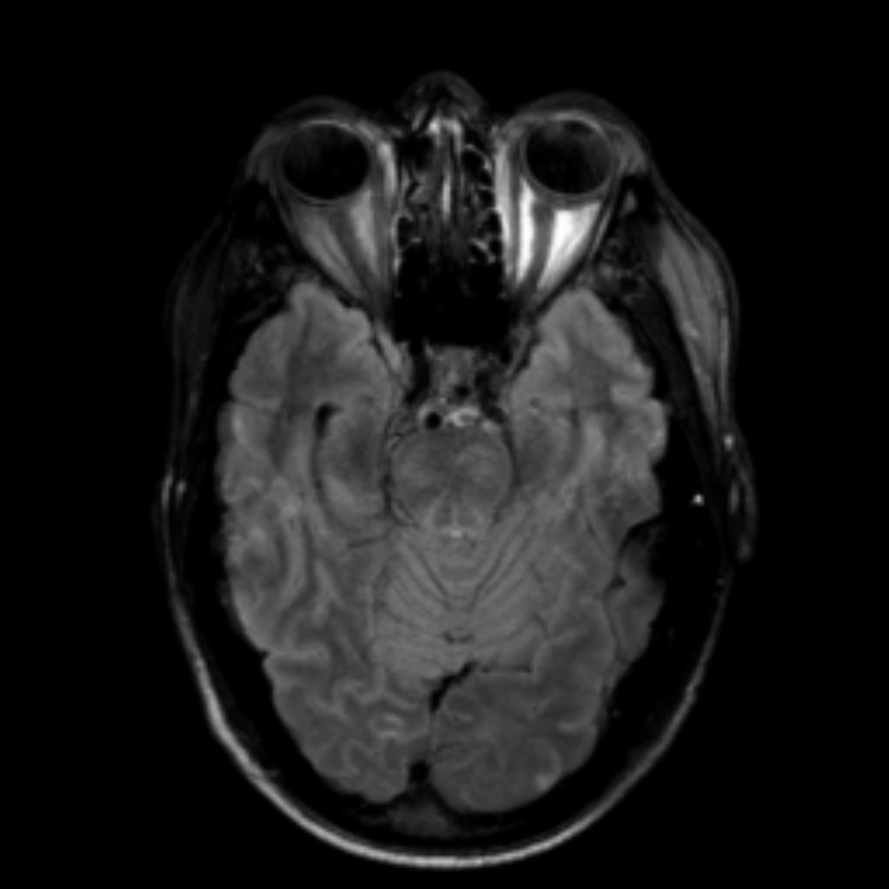}
		\includegraphics[width=0.20\linewidth]{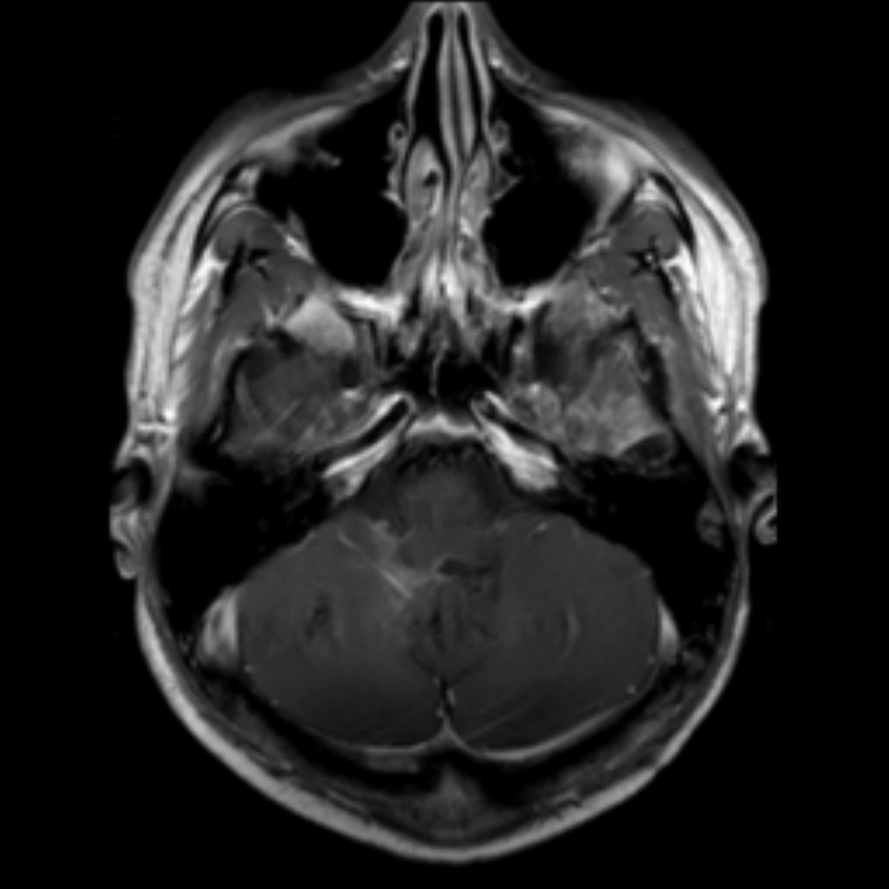}
		\includegraphics[width=0.20\linewidth]{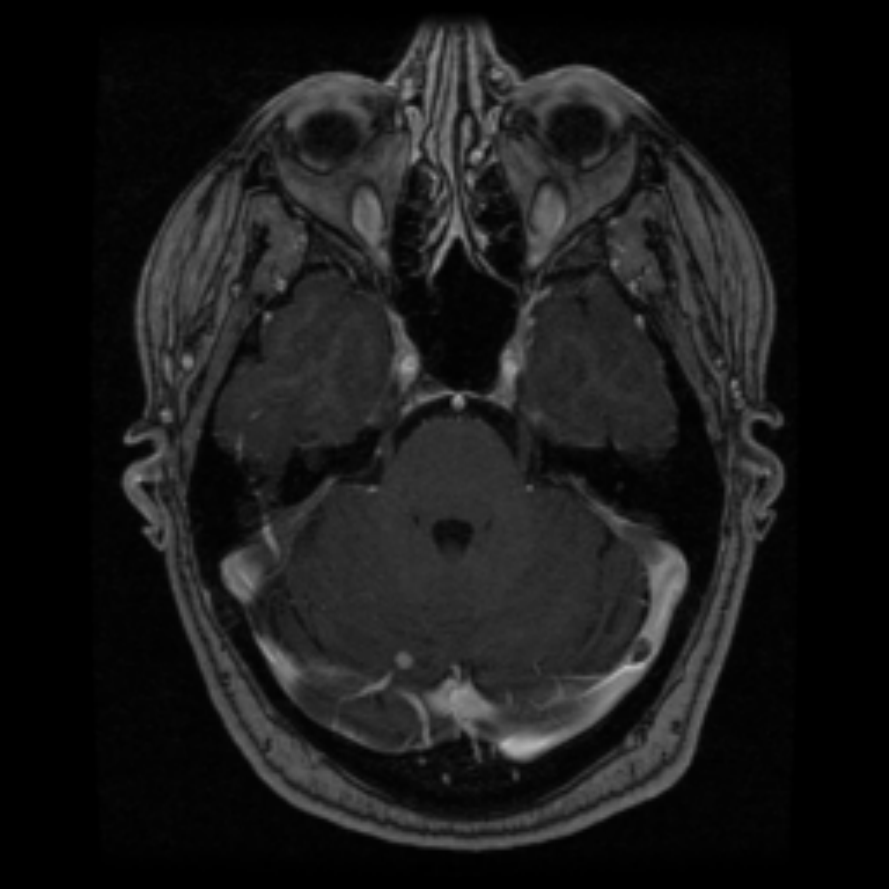}
	\end{center}   
	\caption{Sample images of four LDPO clusters with associated clinically semantic key words, containing the information of (likely appeared) anatomies, pathologies, their attributes and imaging protocols or properties.}  
	\label{fig:Cluster_Sample:png}
\end{figure*}

\begin{figure*}
	\begin{center}
		\includegraphics[width=0.96\linewidth]{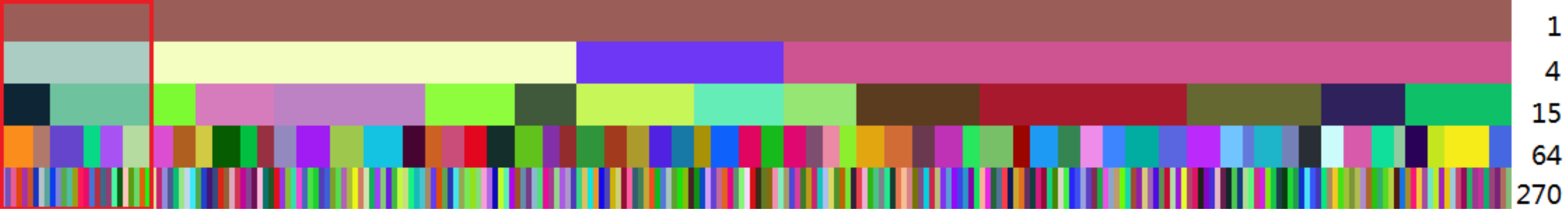}
		\includegraphics[width=0.96\linewidth]{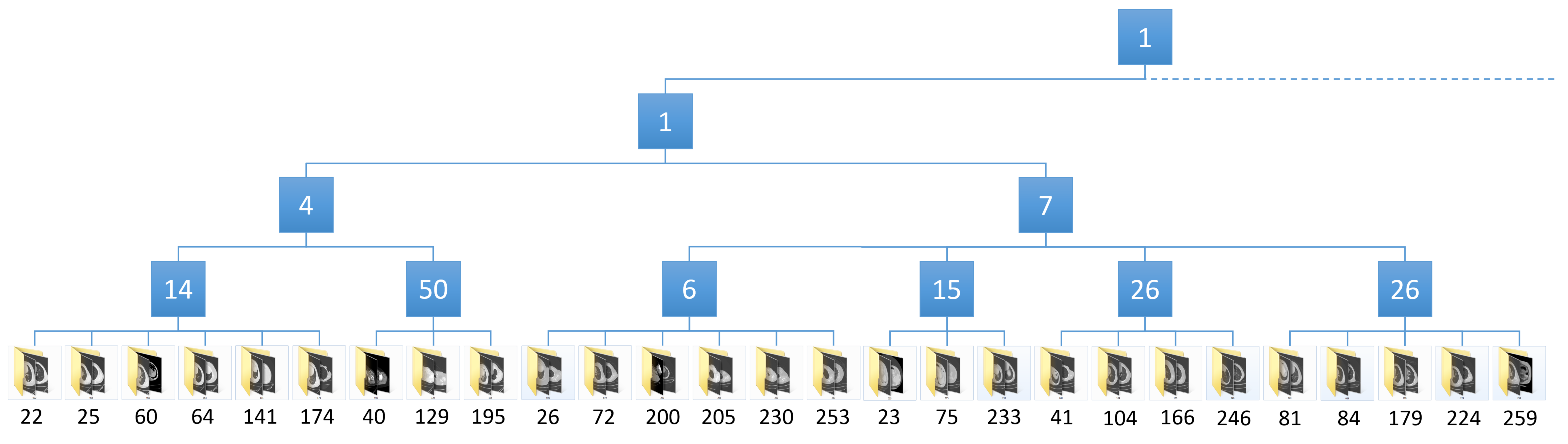}
	\end{center} 
	\caption{Five-level hierarchical categorization is illustrated with a randomized color for each cluster. Sample images and detailed tree structures from a branch (highlighted with a red bounding box) are also shown. {\it The high majority of images in the clusters of this branch are verified as CT Chest scans by radiologists}.} 
	\label{fig:CL_Tree:png}
\end{figure*}

\subsection{LDPO Categorization and Auto-annotation Results}\label{sec-label-results} 

The category discovery clusters employing our LDPO method are found to be more visually coherent and cluster-wise balanced in comparison to the results in \cite{Shin2015} where clusters are formed only from text information ($\sim780K$ radiology reports). Fig.~\ref{fig:Clusters:png} {\bf Left} shows the image numbers for each cluster from the AlexNet-FC7-Topic setting. The numbers are  uniformly distributed with a mean of 778 and standard deviation of 52. Fig.~\ref{fig:Clusters:png} {\bf Right} illustrates the relation of clustering results derived from image cues or text reports~\cite{Shin2015}. Note that there is no instance-balance-per-cluster constraints in the LDPO clustering. The clusters in \cite{Shin2015} are highly uneven: 3 clusters inhabit the majority of images. Fig.~\ref{fig:Cluster_Sample:png} shows sample images and top-10 associated key words from 4 randomly selected clusters (more results in the supplementary material). The LDPO clusters are found to be semantically or clinically related to the corresponding key words, containing the information of (likely appeared) anatomies, pathologies (e.g., adenopathy, mass), their attributes (e.g., bulky, frontal) and imaging protocols or properties. 

Next, from the best performed LDPO models in Table~\ref{tab:CNN-Acc}, {\bf AlexNet-FC7-Topic} has {\bf Top-1} classification accuracy of 0.8109 and {\bf Top-5} accuracy 0.9412 with 270 formed image categories; {\bf AlexNet-FC7-ImageNet} achieves accuracies of 0.8099 and 0.9547, respectively, from 275 discovered classes. In contrast, \cite{Shin2015} reports {\bf Top-1} accuracies of 0.6072, 0.6582 and {\bf Top-5} as 0.9294, 0.9460 on 80 text only computed classes using AlexNet \cite{krizhevsky2012imagenet} or VGGNet-19 \cite{simonyan2014very}, respectively. Markedly better accuracies (especially on {\bf Top-1}) on classifying higher numbers of classes (being generally more difficult) highlight advantageous quality of the LDPO discovered image clusters or labels. This means that the LDPO results have rendered significantly better performance on automatic image labeling than the most related previous work \cite{Shin2015}, under the same radiology database. {\it After the subjective evaluation by two board-certified radiologists, {\bf AlexNet-FC7-Topic} of 270 categories and {\bf AlexNet-FC7-ImageNet} of 275 classes are preferred, out of total six model-encoding setups. Interestingly, both CNN models have no deep feature encoding built-in and preserve the gloss image layouts (capturing somewhat global visual scenes without unordered FV or VLAD encoding schemes \cite{Cimpoi2015Deep,Cimpoi2015Filter,Jegou2012VLAD}.)}. 

\begin{figure}
	\begin{center}
		\includegraphics[width=0.90\linewidth]{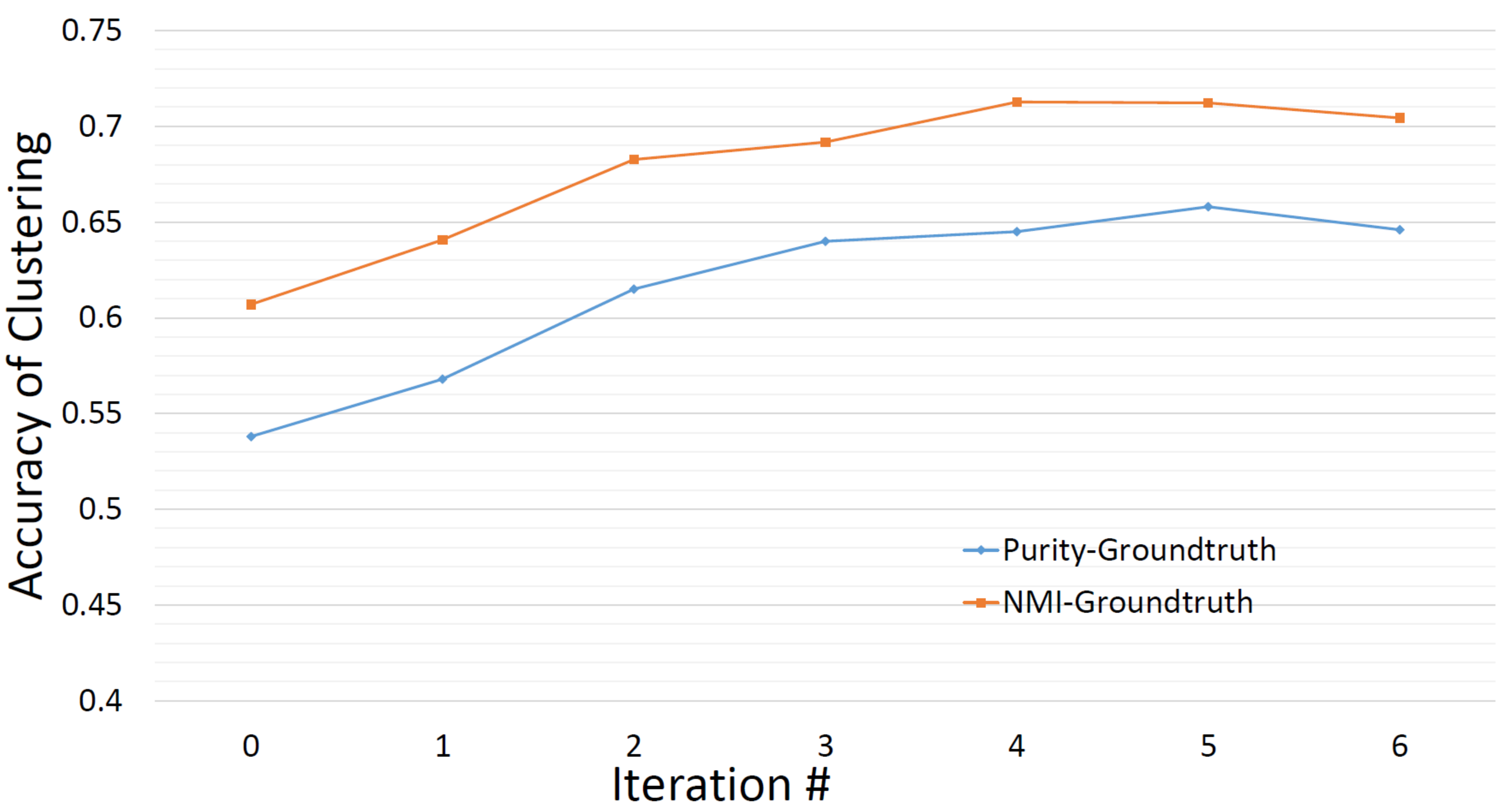}
	\end{center} 
	\caption{Purity (Accuracy) and NMI plots between the ground truth classes and LDPO discovered clusters versus the iteration numbers.} 
	\label{fig:texture:png}
\end{figure}

For the quantitative validation, LDPO is also evaluated on the Texture-25 dataset as an unsupervised texture classification problem. The purity and NMI are computed between the resulted LDPO clusters per iteration and the ground truth clusters (of 25 texture image classes \cite{Dai2015EnProDeepFets,Lazebnik2005Sparse}) where purity becomes classification accuracy. {\bf AlexNet-FC7-ImageNet} is employed and the quantitative results are plotted in Fig. \ref{fig:texture:png}. Using the same clustering method of k-means, the purity or accuracy measurements improve from 53.9\% (0-th) to 66.1\% at the 6-th iteration, indicating that LDPO indeed learns better deep image features and labels in the looped process. Similar trend is found for another texture dataset \cite{Cimpoi2015Filter}. Exploiting LDPO for other domain transfer based auto-annotation tasks will be left as future work.

The final trained CNN classification models allow to compute the pairwise category similarities or affinity scores using the CNN classification confusion values between any pair of classes (Sec. \ref{sec:hcl}). Affinity Propagation algorithm is called recursively to form a hierarchical category tree. The resulted category tree has (270, 64, 15, 4, 1) different class labels from bottom (leaf) to top (root). The random color coded category tree is shown in Fig. \ref{fig:CL_Tree:png}. The high majority of images in the clusters of this branch are verified as CT Chest scans by radiologists. {\it Enabling to construct a semantic and meaningful hierarchy of classes offers another indicator to validate the proposed LDPO category discovery method and results.} Refer to the supplementary material for more results. We will make our trained CNN models, computed deep image features and labels publicly available upon publication.

%

\section{Conclusion \& Future Work} 

In this paper, we present a new Looped Deep Pseudo-task Optimization framework to extract visually more coherent and semantically more meaningful categories from a large scale medical image database. We systematically and extensively conduct experiments under different settings of the LDPO framework to validate and evaluate its quantitative and qualitative performance. The measurable LDPO ``convergence'' makes the ill-posed auto-annotation problem well constrained without the burden of human labeling costs. For future work, we intend to explore the feasibility/performance on implementing our current LDPO clustering component by deep generative density models \cite{Bengio-et-al-2015-Book,Salakhutdinov2015Learning,Kingma2014SSL}. It may therefore be possible that both classification and clustering objectives can be built into a multi-task CNN learning architecture which is ``end-to-end'' trainable by alternating two task/cost layers during SGD optimization \cite{Tzeng2015Simultaneous}. 

\bibliographystyle{ieee}
\bibliography{LDPO_eccv2016}
\end{document}